%% file: CameraReady2027.tex
\documentclass[letterpaper]{article} 
\usepackage{aaai2027}  
\usepackage[hyphens]{url}  
\usepackage{graphicx} 
\usepackage{natbib}  
\usepackage{caption} 
\usepackage{algorithm}
\usepackage{algorithmic}

\usepackage{newfloat}
\usepackage{listings}
\DeclareCaptionStyle{ruled}{labelfont=normalfont,labelsep=colon,strut=off} 
\floatstyle{ruled}
\newfloat{listing}{tb}{lst}{}
\floatname{listing}{Listing}

\usepackage{booktabs}

\usepackage{algorithm}
\usepackage{algorithmic}
 \usepackage{amsmath,amssymb,amsthm,bm}
  \usepackage[dvipsnames]{xcolor}

  \definecolor{ enscol}{HTML}{1F7A8C}   
  \definecolor{protcol}{HTML}{C05621}   
  \definecolor{diffcol}{HTML}{6B46C1}   
  \newcommand{\ens}[1]{\textcolor{ enscol}{#1}}
  \newcommand{\prot}[1]{\textcolor{protcol}{#1}}
  \newcommand{\diff}[1]{\textcolor{diffcol}{#1}}

\usepackage{graphicx}      
\usepackage{booktabs}      
\usepackage{pifont}        
\usepackage[table]{xcolor} 

\newcommand{\methodname}{\textsc{\diff{Ma}\prot{Vi}Seg}}

\definecolor{bestbg}{RGB}{198,219,239}    
\definecolor{secondbg}{RGB}{222,235,247}  
\newcommand{\best}[1]{\cellcolor{bestbg}\textbf{#1}}
\newcommand{\second}[1]{\cellcolor{secondbg}\underline{#1}}
\definecolor{gainbg}{RGB}{198,239,206}
\newcommand{\gain}[1]{{\setlength{\fboxsep}{1.5pt}\colorbox{gainbg}{$#1$}}}
\newcommand{\cmark}{\ding{51}}
\newcommand{\xmark}{\ding{55}}

\usepackage{newfloat}
\usepackage{listings}
\DeclareCaptionStyle{ruled}{labelfont=normalfont,labelsep=colon,strut=off} 
\floatstyle{ruled}
\newfloat{listing}{tb}{lst}{}
\floatname{listing}{Listing}

\usepackage{booktabs}
\nocopyright
\usepackage{lipsum} 
\usepackage{pifont}
\usepackage{todonotes}

\title{\methodname{}: Manifold Propagation and Visual Prototypes for Zero-Shot Open-Vocabulary Segmentation in Diffusion Transformers}
\author{
    Rajatsubhra Chakraborty\textsuperscript{\rm 1}\equalcontrib,
    Xujun Che\textsuperscript{\rm 1}\equalcontrib,
    Ritabrata Chakraborty\textsuperscript{\rm 2},
    Xi Niu\textsuperscript{\rm 1},
    Depeng Xu\textsuperscript{\rm 1}
}
\affiliations{
    \textsuperscript{\rm 1}University of North Carolina at Charlotte, Charlotte, NC, USA\\
    \textsuperscript{\rm 2}Manipal University Jaipur, Jaipur, Rajasthan, India\\
    \{rchakra6, xche, xniu2, dxu7\}@charlotte.edu, ritabrata.229301716@muj.manipal.edu
}
\begin{document}

\maketitle

\begin{abstract}
\input{Sections/0_Abstract}
\end{abstract}

\input{Sections/1_Introduction}
\input{Sections/2_Related_Works}
\input{Sections/4_METHODOLOGY_NEW}
\input{Sections/5_Experiments}
\input{Sections/6_Limitations}
\input{Sections/7_Conclusion}
\bibliography{aaai2027}
\input{Sections/X_supplementary}
\end{document}

%% file: Sections/0_Abstract.tex
Text-to-image diffusion transformers learn about objects and scenes by learning to generate them, making them strong candidates for training-free zero-shot open-vocabulary semantic segmentation. State-of-the-art attribution methods score each pixel independently, comparing its features against a fixed text-derived class representation, whether as an output-space similarity or as a cross-attention weight. This discards structured signals the model itself exposes: the temporal structure of the generative trajectory, the visual appearance statistics of each concept, and the image's own pairwise feature geometry. We present \methodname{}, a training-free refinement layer that recovers these signals. Because its operators consume only a pixel-by-concept score field and a pixel feature space, \methodname{} is capture-agnostic rather than tied to one attribution method. Across six benchmarks it achieves the strongest overall results among training-free methods, including the best mIoU on every benchmark. Interestingly, gains are largest where the initial capture is weakest, and individual operators contribute depending on the noise in the field they refine. Our results indicate that diffusion transformers carry more concept-level information than current attribution methods recover, and that much of it is lost on the way to the mask rather than absent from the model.

%% file: Sections/1_Introduction.tex
\section{Introduction}
\label{sec:intro}

\begin{figure}[t]
    \centering
    \includegraphics[width=\linewidth]{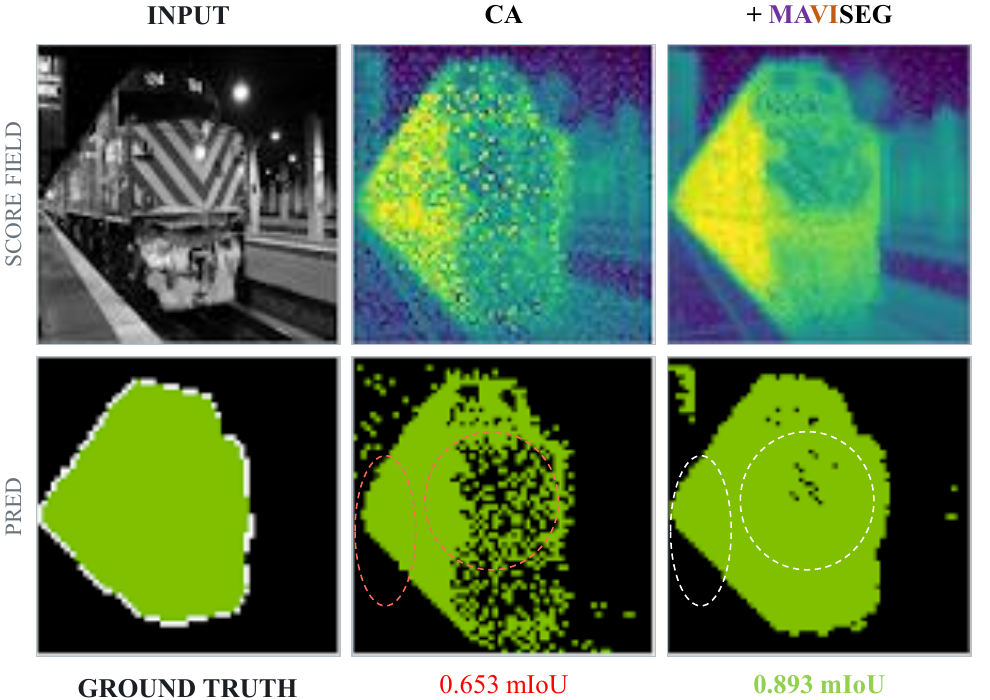}
    \caption{\textbf{Diffusion transformers already know where things are.}
Existing methods like ConceptAttention (CA) score each token against one text
vector at a single timestep, leaving hollow interiors and bleed across edges
(red). \methodname{} recovers the \ens{temporal}, \prot{appearance} and
\diff{geometric} signals the same forward pass exposes, improving attribution
maps with no retraining (white).}
    \label{fig:teaser}
\end{figure}

To paint a convincing dog on a sofa, a text-to-image model has to decide which
pixels are dog and which are sofa. It cannot render the scene otherwise.
Localisation is therefore not something these models were taught but something
they must already possess in order to generate at all, which makes a frozen
generator an unusual kind of segmenter, one that never saw a mask yet knows
where things are.

This property is what makes generative backbones attractive for
open-vocabulary semantic segmentation (OVSS), where a user supplies an
arbitrary set of concepts in words at test time and the model must label every
pixel against them without having seen a single segmentation mask during
training. Early work read that knowledge out of the cross-attention of U-Net
diffusion models \citep{rombach2022ldm,tang2023daam}, but the shift to
Multimodal Diffusion Transformers (MM-DiT) such as Stable Diffusion~3
\citep{esser2024sd3} and FLUX \citep{flux2024} exposed a far cleaner signal. In
these backbones image and text tokens interact through joint attention, and a
key recent finding is that the output (value) space of MM-DiT layers is
linearly concept-separable, so projecting an image token onto a text-derived
concept direction yields a sharp saliency map. ConceptAttention
\citep{helbling2025conceptattention} turns this into state-of-the-art zero-shot
segmentation without any training, and independent read-outs of the same
backbones \citep{kim2025seg4diff} confirm the pattern. A frozen generator is
now a competitive open-vocabulary segmenter.

Which raises an awkward question. If the backbone already knows, why are the
masks still wrong in such recognisable ways (Fig.~\ref{fig:teaser})?
Our answer is that the loss happens after the backbone, in the read-out.
Current methods classify each token in isolation, taking a single dot product
between the token's vector and one fixed text-derived concept vector, per
concept, thresholded independently everywhere. Three structured signals that
the very same forward pass makes available are simply discarded. The first is
the \ens{temporal axis}, since the generative trajectory forms an image's
semantics over many timesteps while the read-out is taken at just one. The
second is \prot{appearance}, since the concept anchor is the word ``dog''
rather than what dogs look like, even though the model's own feature space
clusters instances of a class tightly. The third is \diff{geometry}, since
tokens are judged one at a time with no use of the image's own token-to-token
similarity structure. Each discarded signal leaves its own fingerprint, namely
hollow object interiors where boundaries are confident but centres are not,
concept bleeding across edges, and confusions between concepts whose words sit
close together but whose appearances plainly do not. The concept-level evidence
needed to avoid these errors is present in the backbone and is lost in transit
to the mask.

\methodname{} puts each signal back at the read-out, leaving the capture
untouched. For the \ens{temporal axis} it reads the score field at several
points along the trajectory instead of one and averages them, which turns a
choice of timestep into an aggregate over timesteps. For \prot{appearance} it
adds a second, visually grounded scoring channel whose per-concept anchors are
harvested from unlabelled images by the zero-shot scorer itself, so the word
``dog'' is no longer the only thing a dog token is compared against. For
\diff{geometry} it lets confident tokens inform ambiguous ones through a
self-tuning random walk on the image's own feature graph, which fills interiors
without blurring the boundaries. Each operator consumes only a token-by-concept
score field and a token feature space, the two objects every attribution method
already produces, so the same stack refines captures that read entirely
different internal signals in different backbones.

Overall, our contributions are as follows:
\begin{enumerate}
  \item  We show that MM-DiT
  attribution is limited by its read-out rather than by its backbone, since
  scoring every token independently against one text vector at one timestep
  discards the \ens{temporal}, \prot{appearance} and \diff{geometric} structure
  the same forward pass already exposes.

  \item We introduce \methodname{}, a training-free refinement layer with one
  operator per discarded signal, \ens{timestep-diversity ensembling},
  \prot{bootstrapped visual prototypes} and \diff{manifold diffusion}, using no
  masks, no auxiliary encoder and no parameter updates.

  \item  \methodname{} attains
  the best mIoU on every benchmark among training-free methods, improving
  ConceptAttention by $+4.0$ mIoU and Seg4Diff by $+4.5$ mIoU on average, with
  the largest gains where the underlying capture is weakest.

\end{enumerate}


%% file: Sections/2_Related_Works.tex
\section{Related Work}
\label{sec:related}
 
\subsection{Open-Vocabulary Semantic Segmentation}
Open-vocabulary semantic segmentation (OVSS)\footnote{OVSS and OVS have been used interchangably throughout literature, both refer to the same task.} has emerged as an important downstream task to assess vision capabilities of models.
Discriminative backbones such as CLIP \citep{radford2021clip} have been initially studied, whose
contrastive vision-language alignment supplies the open-set vocabulary.  
One family trains a segmentation head on top of that alignment, either by aligning pixel embeddings to text directly \citep{li2022lseg,xu2022groupvit} or by first proposing regions and then classifying them with a frozen vision--language model \citep{ghiasi2022openseg,liang2023ovseg,xu2023san,cho2024catseg}. These methods improve accuracy, but they pay for it in mask or caption supervision and in training cost. A second family is training-free and works instead on the fact that CLIP's patch tokens are semantically correct but spatially blurred. The dominant remedy is attention surgery: replacing or removing the query--key mixing of the final block so that patches stop attending globally \citep{zhou2022maskclip,wang2024sclip,bousselham2024gem,lan2024clearclip,hajimiri2025naclip}. Interestingly, some works accept CLIP's granularity as a hard ceiling and imports spatial structure from a second frozen model, using DINO \citep{dino} or SAM \cite{sam} features to re-aggregate CLIP tokens \citep{lan2024proxyclip,wysoczanska2024clipdinoiser,shi2025trident}. The latter aligns with our motivation that the geometry has to be supplied by a model other than the one that holds the vocabulary. Text-to-image generators (discussed below) are relevant here precisely because they hold both at once.
 
\subsection{Diffusion Models for OVSS}
WIth the rise of text-to-image (T2I) models \cite{ldm,dalle,gabeur2026imagegeneratorsgeneralistvision}, these models were used for OVSS. Some methods use them as a
data engine of sorts, such as DiffuMask \citep{wu2023diffumask} and related work which synthesise
images together with pixel labels to train a downstream segmenter. This is
orthogonal to our aim, since a segmenter still has to be trained, whereas we are motivated towards training-free methods. Some methods read segmentation out of the model's own
signals. Initial frameworks \citep{tang2023daam,marcosmanchon2024ovam} exploited U-Net cross-attention, where each concept word
attends to the pixels it generates. ODISE \citep{xu2023odise} couples internal diffusion features with a
mask generator for open-vocabulary panoptic segmentation; and
EmerDiff \citep{namekata2024emerdiff} shows that pixel-level semantics emerge in diffusion features even
without explicit attention read-out . A prototype-based
branch, OVDiff \citep{karazija2024ovdiff} and FreeDA \citep{barsellotti2024freeda},
generates a support set of images per queried class and pools features over
them to form class prototypes, which are then matched against the test image. Moving towards diffusion transformers (DiTs), Concept-Attention \citep{helbling2025conceptattention} shows the output space of
MM-DiT layers is linearly concept-separable, while Seg4Diff \citep{kim2025seg4diff} obtains a
comparable signal from image-to-text attention, finding expert layers for OVSS. A handful of methods pair such a
capture with a refinement stage, such as the multi-resolution
cross-attention fusion of DiffSegmenter \citep{wang2023diffsegmenter} and the
iterated self-attention propagation of iSeg \citep{sun2024iseg}.

\paragraph{Our Work.}\methodname{} sits with the training-free refinement methods that improve OVSS on
generative backbones, but unlike DiffSegmenter, iSeg and DiffSeg, which bind to one cross-attention capture, and unlike the
OVDiff/FreeDA prototype pipelines, which need a separate
generation-plus-DINO/CLIP stage, it consumes only a score field and a feature
space and harvests prototypes from unlabelled real images with the zero-shot
scorer itself. 

%% file: Sections/4_METHODOLOGY_NEW.tex
\begin{figure*}
    \centering
    \includegraphics[width=\textwidth]{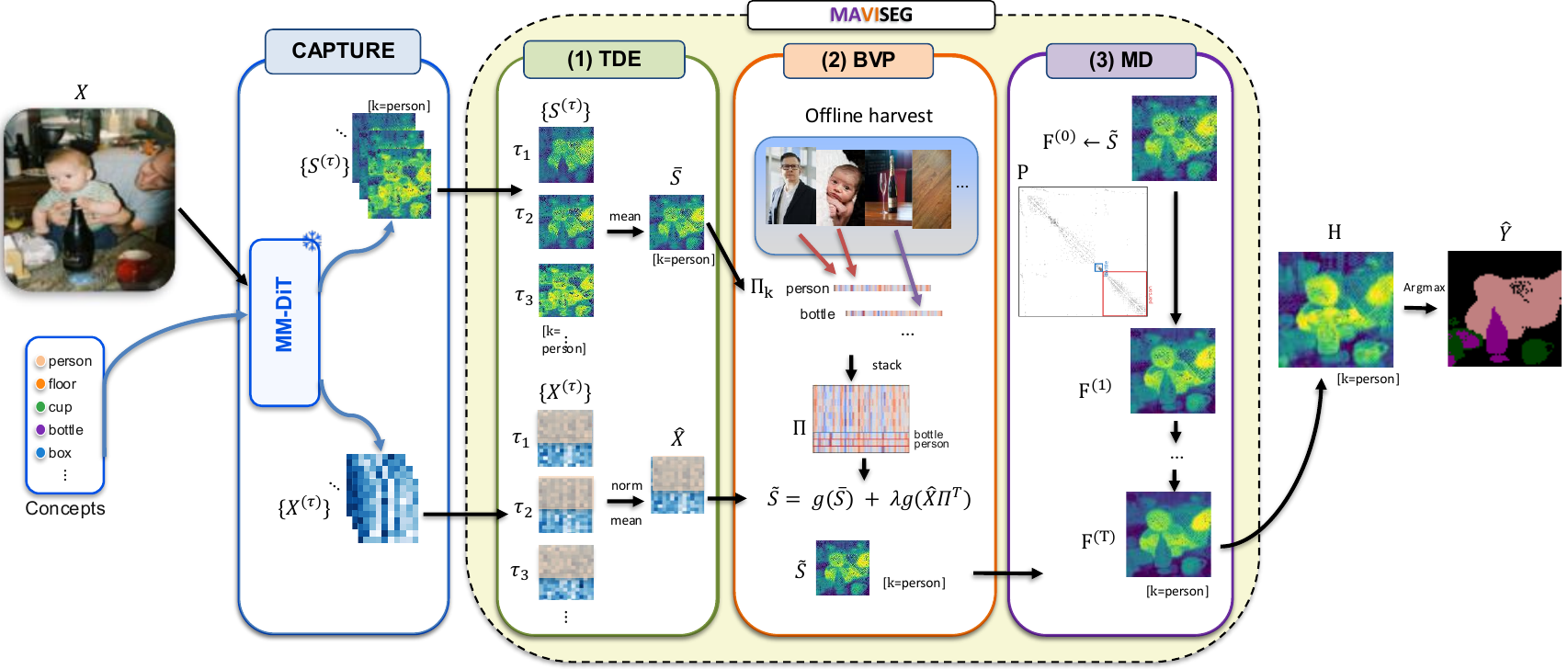}
    \caption{\textbf{\methodname{} architecture overview. }(1) \ens{\textbf{TDE}} averages captures for $x$ over a small set of timesteps $\{\mathbf{S}^{(\tau)},\mathbf{X}^{(\tau)}\}_{\tau\in\mathcal{T}}$  into ${(\bar{\mathbf{S}},\hat{\mathbf{X}})}$.
(2) \prot{\textbf{BVP}} harvests the prototype bank $\Pi$
  from unlabelled images $\mathcal{U}$ and adds per-concept anchor scores to the field. (3) \diff{\textbf{MD}} propagates $\mathbf{F}$ with a confined transition matrix $\mathbf{P}$. We visualize the score field of class "person" in $S,F,H$. The token positions in $\hat{X}, P$ corresponding to the groundtruth class are highlighted for visualization but not used in the algorithm.}
    \label{fig:architecture}
\end{figure*}

\section{Methodology}
\label{sec:method}

\subsection{Problem Setup}
\label{sec:problem_setup}

\paragraph{Task.}
Given an image $x$ and a concept set
$\mathcal{C}=\{c_1,\ldots,c_K\}$ supplied at test time, open-vocabulary
semantic segmentation (OVSS) assigns every location in $x$ one of the $K$ labels. We study OVSS in zero-shot, hence the backbone is a frozen
text-to-image diffusion transformer, no masks, boxes, or captions are used at
any stage, and no parameters are updated\footnote{Since pixel predictions are made on the model's
latent token grid of $N$ positions and upsampled to the evaluation resolution, we refer to tokens rather than pixels.}.

\paragraph{Captures.}
A diffusion attribution method can provide spatial concept
evidence. For some methods, the forward pass compares image-token outputs against
text-derived concept vectors, others read the attention between image and text
tokens. However, they expose the same two objects, so we describe them through a
single \underline{capture} interface. At diffusion timestep $\tau$, a capture $\Phi$
produces
\begin{equation}
\Phi_{\tau}(x,\mathcal{C})
=
\left(
    \mathbf{S}^{(\tau)},
    \mathbf{X}^{(\tau)}
\right),
\mathbf{S}^{(\tau)}\in\mathbb{R}^{N\times K}, 
\mathbf{X}^{(\tau)}\in\mathbb{R}^{N\times d},
\label{eq:capture}
\end{equation}
where $\mathbf{S}^{(\tau)}_{pk}$ is the affinity between token $p$ and concept
$c_k$, and $\mathbf{X}^{(\tau)}_{p}$ is the representation of token $p$ in the
feature space of the frozen backbone. We call $\mathbf{S}$ the \emph{score
field} and $\mathbf{X}$ the \emph{token features}. Because the interface does
not depend on how the scores are obtained, it covers output-space and
attention-based attribution alike.

\paragraph{Conventional Read-out.}
A conventional method captures at one timestep $\tau_0$ and labels each token
by its highest-scoring concept,
\begin{equation}
    \hat{y}^{\,\mathrm{base}}_{p}
    =
    \arg\max_{k\in\{1,\ldots,K\}}
    \mathbf{S}^{(\tau_0)}_{pk},
    \label{eq:baseline_readout}
\end{equation}
The predicted mask therefore depends on the
backbone only through one score matrix, and every token is decided in isolation
from every other.

\subsection{Overview}

The conventional read-out, Eq.~\eqref{eq:baseline_readout}, assigns each token
its highest-scoring concept in a single matrix $\mathbf{S}^{(\tau_0)}$, and thereby
discards three signals the same forward pass exposes. It reads one point of a
trajectory spanning many timesteps, ignoring the \ens{temporal} axis. It scores
tokens only against text-derived concept vectors, ignoring \prot{appearance},
which $\mathbf{X}$ encodes. It decides each token independently, ignoring the
\diff{geometry} of $\mathbf{X}$ that relates tokens to one another. The
corresponding failures are visible in the output: predictions that shift with
the choice of $\tau_0$, confusions between concepts whose words are close but
whose appearances are not, and hollow interiors with concept bleed across edges.

\paragraph{Our Scope.}
We keep $\Phi$ fixed and ask instead how much of the segmentation quality is
recoverable at the decoding stage.  Let
\begin{equation}
    \mathcal{Z}_{x}
    =
    \left\{
        \left(
            \mathbf{S}^{(\tau)},
            \mathbf{X}^{(\tau)}
        \right)
    \right\}_{\tau\in\mathcal{T}}
    \label{eq:capture_set}
\end{equation}
be the captures collected for $x$ over a small set of timesteps $\mathcal{T}$.
We study refinement operators
\begin{equation}
    \mathbf{H}
    =
    \mathcal{R}\!\left(\mathcal{Z}_{x};\,\mathbf{\Pi}\right),
    \qquad
    \hat{y}_{p}
    =
    \arg\max_{k}\mathbf{H}_{pk},
    \label{eq:refinement_objective}
\end{equation}
which return a refined score field $\mathbf{H}\in\mathbb{R}^{N\times K}$ using
no segmentation annotations, no parameter updates, and no access to the
internals of $\Phi$. Here $\mathbf{\Pi}$ is a visual prototype bank precomputed
once from the captures of an unlabelled image pool by the same frozen $\Phi$
(\S\ref{sec:prototypes}); it is the only cross-image input.
We recover these signals at the decoding stage, leaving the capture untouched,
by instantiating the refinement operator $\mathcal{R}$ of
Eq.~\eqref{eq:refinement_objective}, which maps the captures $\mathcal{Z}_x$ of
Eq.~\eqref{eq:capture_set} to a refined field $\mathbf{H}$ read out in place of
$\mathbf{S}^{(\tau_0)}$. Since $\mathcal{R}$ consumes only $\mathbf{S}$ and $\mathbf{X}$, it
is independent of how they were produced.

\paragraph{Our Architecture.}
We present \methodname{}, a training-free refinement layer that recovers these signals. Because its operators consume only a pixel-by-concept score field and a pixel feature space, \methodname{} is capture-agnostic rather than tied to one attribution method. Fig. \ref{fig:architecture} shows the overall architecture of \methodname{}. It contains three operators,  \ens{Timestep-Diversity Ensemble (TDE)}, \prot{Bootstrapped Visual Prototypes (BVP)}, and \diff{Manifold Diffusion (MD)}.
\ens{\textbf{TDE}} recovers the temporal signals $\{\mathbf{S}^{(\tau)},\mathbf{X}^{(\tau)}\}_{\tau\in\mathcal{T}}$ along the generative  trajectory $\mathcal{T}$ and averages the score field and token features into ${(\bar{\mathbf{S}},\hat{\mathbf{X}})}$. 
\prot{\textbf{BVP}} recover appearance by
  adding per-concept anchor scores against the prototype bank $\Pi$
  harvested from unlabelled images $\mathcal{U}$. \diff{\textbf{MD}} recovers geometry through a
  self-tuning random walk on the image's own feature graph. 
It propagates the dual-
channel field $\mathbf{F}$ by inverse-free power iteration with a confined transition matrix $\mathbf{P}$.
The final refined field $\mathbf{H}$ is used for the read-out. The steps are shown in Algorithm \ref{alg:maviseg}.

\begin{algorithm}[t]
\caption{\textbf{\methodname{} refinement at inference.} The prototype bank $\mathbf{\Pi}$
is harvested once, offline, using Eq.~\eqref{eq:harvest}. 
}
\label{alg:maviseg}
\begin{algorithmic}[1]
\REQUIRE captures $\{\mathbf{S}^{(\tau)},\mathbf{X}^{(\tau)}\}_{\tau\in\mathcal{T}}$,
prototype bank $\mathbf{\Pi}$, scalars $\lambda,\kappa,m,\alpha,\rho,T_{\mathrm{it}}$
\ENSURE per-token label map $\hat{\mathbf{y}}$
\STATE $\bar{\mathbf{S}}\leftarrow \frac{1}{|\mathcal{T}|}\sum_{\tau}\mathbf{S}^{(\tau)}$,
$\bar{\mathbf{X}}\leftarrow \frac{1}{|\mathcal{T}|}\sum_{\tau}\mathbf{X}^{(\tau)}$
\COMMENT{\ens{TDE}}
\label{line:ensemble}
\STATE $\hat{\mathbf{x}}_p\leftarrow\bar{\mathbf{X}}[p]/\lVert\bar{\mathbf{X}}[p]\rVert_2$
for all $p$ and stack as $\hat{\mathbf{X}}$
\label{line:normalize_features}
\STATE $\tilde{\mathbf{S}}\leftarrow g(\bar{\mathbf{S}})+\lambda\,g(\hat{\mathbf{X}}\mathbf{\Pi}^{\top})$
\COMMENT{\prot{BVP}}
\label{line:dual_channel}
\STATE construct $\mathbf{W}$ from $\{\hat{\mathbf{x}}_p\}$ using Eq.~\eqref{eq:affinity}
\label{line:graph}
\STATE $\mathbf{D}_{pp}\leftarrow\sum_q\mathbf{W}[p,q]$ and
$\mathbf{P}\leftarrow\mathbf{D}^{-1/2}\mathbf{W}\mathbf{D}^{-1/2}$
\label{line:transition}
\STATE initialize $\mathbf{F}^{(0)}\leftarrow\tilde{\mathbf{S}}$ and iterate
$\mathbf{F}^{(t+1)}\leftarrow\alpha\mathbf{P}\mathbf{F}^{(t)}+(1-\alpha)\tilde{\mathbf{S}}$
for $T_{\mathrm{it}}$ iterations
\COMMENT{\diff{MD}}
\label{line:propagate}
\STATE $\mathbf{H}\leftarrow\rho\mathbf{F}+(1-\rho)\tilde{\mathbf{S}}$
\label{line:blend}
\STATE \textbf{return} $\hat{y}_p \leftarrow \arg\max_k \mathbf{H}[p,k],\ \forall\, p$
\label{line:readout}
\end{algorithmic}
\end{algorithm}

\subsection{\ens{Timestep-Diversity Ensemble.}}
\label{sec:ensemble}

The generative process is non-stationary, so what a capture exposes depends on
where it is read, with early steps carrying coarse layout and later steps
interior detail. The same instability appears across depth, where reading the
identical capture from each deep block in isolation swings the resulting mIoU
sharply from one block to the next (Fig.~\ref{fig:layerwise}). Fixing $\tau_0$
therefore discards evidence and introduces a backbone-dependent hyperparameter.
TDE removes the choice by averaging both objects of the
interface over a small set $\mathcal{T}$ of interior timesteps  (Line~\ref{line:ensemble} in Algorithm~\ref{alg:maviseg}), the endpoints
being either noise-dominated or nearly collapsed,
\begin{equation}
\ens{\bar{\mathbf{S}}}=\frac{1}{|\mathcal{T}|}\sum_{\tau\in\mathcal{T}}\mathbf{S}^{(\tau)},
\qquad
\ens{\bar{\mathbf{X}}}=\frac{1}{|\mathcal{T}|}\sum_{\tau\in\mathcal{T}}\mathbf{X}^{(\tau)},
\label{eq:ensemble}
\end{equation}
after which it  $\ell_2$-normalizes the averaged
features into unit vectors $\hat{\mathbf{x}}_p\leftarrow\bar{\mathbf{X}}[p]/\lVert\bar{\mathbf{X}}[p]\rVert_2$, stacked as the rows of
$\hat{\mathbf{X}}$ (Line~\ref{line:normalize_features} in Algorithm~\ref{alg:maviseg}). 

\subsection{\prot{Bootstrapped Visual Prototypes.}}
\label{sec:prototypes}

Every entry of $\bar{\mathbf{S}}$ is an affinity to a \emph{text} anchor, so
concepts whose words embed close together are scored against near-parallel
directions however differently they appear. This is a failure of calibration
between concepts rather than of localization: a capture can place a horse
correctly in the image and still score it as a cow. What is missing is an anchor
built from appearance, which the prototype pipelines of OVDiff
\citep{karazija2024ovdiff} and FreeDA \citep{barsellotti2024freeda} obtain by
generating a support set per class and encoding it with a separate DINO or CLIP
model. We need neither, since $\hat{\mathbf{X}}$ already clusters instances of a
class tightly and the text channel, though poorly calibrated across concepts, is
a good enough scorer to find confident exemplars of each.

A single offline pass over an unlabelled pool $\mathcal{U}$ therefore suffices.
Because raw MM-DiT inner products vary across images and channels, we first
define a normalizer $g(*) $ that subtracts a score matrix's global mean and
divides it by its median per-token top-two decision margin, a global affine map
that leaves every per-token $\arg\max$ unchanged, i.e., $g(*) = \frac{* - \mu(*)}{\operatorname{median}_p\big(*[p]_{(1)} - *[p]_{(2)}\big)}$, 
where $\mu(*)$ is the global mean and $*[p]_{(1)},*[p]_{(2)}$ denote the largest and
second-largest entry of row $p$. 
Calibrating the normalized scores into distributions
$\mathbf{p}_u=\operatorname{softmax}(g(\bar{\mathbf{S}}_u)/T_{\mathrm{cal}})$ at
temperature $T_{\mathrm{cal}}$, and letting $\mathcal{I}_{u,k}$ index the top
$r\%$ of tokens for concept $c_k$ in each image $u \in \mathcal{U}$, the prototype bank
$\mathbf{\Pi}\in\mathbb{R}^{K\times d}$ stacks
\begin{equation}
\prot{\mathbf{\Pi}_k}=\frac{1}{rN|\mathcal{U}|}
\left(\sum_{u\in\mathcal{U}}\sum_{p\in\mathcal{I}_{u,k}}\hat{\mathbf{x}}^{\,u}_p\right).
\label{eq:harvest}
\end{equation}
Eq.~\eqref{eq:harvest} reads no masks, generates no images and updates no
parameters, so the scorer bootstraps its own appearance anchors from the frozen
model without leaving the zero-shot setting. It is run once per vocabulary and
enters Algorithm~\ref{alg:maviseg} as an input.

BVP then scores every token against every prototype and
adds the result as a second channel (Line~\ref{line:dual_channel} in Algorithm~\ref{alg:maviseg}),
\begin{equation}
\prot{\tilde{\mathbf{S}}}=g(\bar{\mathbf{S}})+\lambda\,g\!\left(\hat{\mathbf{X}}\mathbf{\Pi}^{\top}\right),
\label{eq:dual}
\end{equation}
where passing both terms through $g$ puts them on a common scale of one median
decision margin, making $\lambda$ a meaningful trade-off rather than an
arbitrary scale factor. The channels are complementary, as the text term
generalizes to concepts never harvested while the prototype term separates those
the text term cannot, and the latter reuses the features of
Eq.~\eqref{eq:ensemble} at no additional forward pass. 

\subsection{\diff{Manifold Diffusion.}}
\label{sec:diffusion}

The errors surviving Eq.~\eqref{eq:dual} are errors of \emph{extent} rather than
identity. A token in a hollow interior is ambiguous while its neighbours in
feature space are confident and correct, yet the per-token $\arg\max$ gives the
second no way to inform the first. We therefore require the refined field to
vary smoothly over the feature manifold rather than the image plane, since
spatial smoothing is what blurs the boundaries we want to keep. With
$\delta_{pq}=1-\langle\hat{\mathbf{x}}_p,\hat{\mathbf{x}}_q\rangle$ the cosine distance
between tokens $p$ and $q$, $\mathrm{kNN}_{\kappa}(p)$ the $\kappa$ nearest
neighbours of $p$ under $\delta$, and $\sigma_p$ the distance from $p$ to its
$m$th nearest neighbour with $m\leq\kappa$, We build (Line~\ref{line:graph} in Algorithm~\ref{alg:maviseg})
\begin{equation}
\mathbf{W}[p,q]=
\begin{cases}
\exp\!\left(-\dfrac{\delta_{pq}^{2}}{\sigma_p\sigma_q}\right),
& q\in\mathrm{kNN}_{\kappa}(p),\\[6pt]
0, & \text{otherwise},
\end{cases}
\label{eq:affinity}
\end{equation}
symmetrised by $\mathbf{W}\leftarrow\max(\mathbf{W},\mathbf{W}^{\top})$. Sparsity is what
separates this operator from smoothing, since a dense kernel leaks mass across
semantic discontinuities at a rate set by one global bandwidth, whereas
truncating to $\kappa$ neighbours severs those paths and confines propagation to
the component an object occupies in feature space, while the local bandwidths
$\sigma_p\sigma_q$ let the kernel adapt to regions of differing feature density.

With the degree matrix $\mathbf{D}_{pp}=\sum_q\mathbf{W}[p,q]$,
we form the symmetric normalized transition
$\mathbf{P}=\mathbf{D}^{-1/2}\mathbf{W}\mathbf{D}^{-1/2}$ (Line~\ref{line:transition} in Algorithm~\ref{alg:maviseg}), which has $O(\kappa N)$ non-zeros, and
propagate the dual-channel field by inverse-free power
iteration~\citep{zhou2003consistency,tong2006rwr} (Line~\ref{line:propagate} in Algorithm~\ref{alg:maviseg}),
\begin{equation}
\mathbf{F}^{(t+1)}=\alpha\mathbf{P}\mathbf{F}^{(t)}+(1-\alpha)\tilde{\mathbf{S}},
\label{eq:power_iteration}
\end{equation}
where $\mathbf{F}$ is the propagated field, initialized at $\tilde{\mathbf{S}}$, and
$0<\alpha<1$ is the propagation weight (equivalently, $1-\alpha$ is the restart
weight), for which the iteration converges geometrically without forming an
inverse. We then retain a
fraction $1-\rho$ of the unpropagated score (Line~\ref{line:blend} in Algorithm~\ref{alg:maviseg}),
\begin{equation}
\diff{\mathbf{H}}=\rho\mathbf{F}+(1-\rho)\tilde{\mathbf{S}},
\label{eq:blend}
\end{equation}
so propagation cannot erase the original evidence. 


%% file: Sections/5_Experiments.tex

\section{Experiments}
\label{sec:experiments}


\subsection{Settings}

\paragraph{Datasets.}
We evaluate OVSS on six benchmarks: PASCAL~VOC~2012 \citep{everingham2015voc},
Pascal-Context-59 \citep{mottaghi2014pascalcontext}, ADE20K
\citep{zhou2017ade20k}, Cityscapes \citep{cordts2016cityscapes}, COCO-Object
and COCO-Stuff-27 \citep{lin2014coco,caesar2018cocostuff}; per-dataset
statistics are given in Appendix~\ref{supp:datasets}. Together they span object-centric
scenes, cluttered indoor and outdoor contexts, urban street imagery;
vocabularies from $19$ to $150$ classes and both coarse and fine-grained concept regimes.

\paragraph{Metrics.}
Following recent works
\citep{helbling2025conceptattention,kim2025seg4diff}, we report
mean intersection-over-union (mIoU), pixel accuracy (pAcc) and mean average
precision (mAP). Since mAP scores the soft field before the arg-max,
it reflects ranking rather than the final decision, so we treat mIoU and pAcc
as the primary metrics.

\paragraph{Implementation details.}
Unless stated otherwise, our main results use the FLUX-schnell backbone
\citep{flux2024} with $4$ inference steps and guidance $0$, capturing from deep
MM-DiT layers $10$--$18$ at timesteps $\mathcal{T}=\{1,2,3\}$. For the Seg4Diff
capture we use the corresponding noise levels $\{4,8,12\}$. The refinement
operators use a residual blend $\rho=0.98$, the share of the diffused field
retained at read-out, and propagation weight $\alpha=0.9$, equivalently a
restart weight $1-\alpha$, for manifold diffusion. The feature graph keeps
$\kappa=60$ neighbours per token with the local bandwidth set by the $m=7$-th
neighbour. The prototype channel uses a blend $\lambda=0.5$ against the text
channel, a harvest fraction $r=2\%$ of tokens per concept per image, a pool of
$|\mathcal{U}|=300$ unlabelled images, and a calibration temperature $T=0.3$.
We run $T_{\mathrm{it}}=30$ diffusion iterations, at which point the geometric
iteration has converged. Prototypes are
harvested once per dataset from unlabelled training images disjoint from the
evaluation set, hence \methodname{} remains fully inductive. Exact commands,
concept strings and library versions are in Appendix~\ref{supp:impl}.

\paragraph{Baselines.}
We compare against training-free methods from two backbone families, matching
the grouping in Table~\ref{tab:main}. On the \emph{discriminative} side, we
evaluate three attribution read-outs of a single CLIP ViT-H-14
\citep{radford2021clip,ilharco2021openclip}: Chefer relevance propagation
\citep{chefer2021transinterp}, text-conditioned GradCAM
\citep{selvaraju2017gradcam}, and the TextSpan decomposition
\citep{gandelsman2024textspan}. On the \emph{generative} side, we evaluate four
diffusion attribution methods, each on the backbone of its original
publication: DAAM on SDXL \citep{tang2023daam,podell2023sdxl}, OVAM on SD1.5
\citep{marcosmanchon2024ovam}, Seg4Diff on SD3-medium \citep{kim2025seg4diff},
and ConceptAttention on FLUX \citep{helbling2025conceptattention}. We also
include reproduced results for DiffSegmenter \citep{wang2023diffsegmenter}, a comparable baseline
that already couples a capture with a refinement stage.

\input{Tables/main_results}

\subsection{Comparison with the State of the Art}
\paragraph{OVSS performance.}Table~\ref{tab:main} reports mIoU, pixel accuracy and mAP for all baselines and
for \methodname{} across the six benchmarks. For discriminative methods, we observe an early plateau in performance. Particularly TextSpan (strongest method) reaches $0.434$ mIoU on VOC
against $0.527$ for a raw generative capture, and the gap persists across
PC-59 and COCO-Stuff-27. CLIP's contrastive alignment supplies the vocabulary
but not the granularity, since its patch tokens are semantically correct and
spatially blurred, and no read-out of those tokens recovers what the
representation never localised. Among the generative methods, the U-Net-based
captures are limited by calibration rather than localisation. For example, DAAM attains
$0.757$ mAP on VOC while collapsing to $0.059$ mIoU, its per-concept maps
individually informative but not comparable across concepts, and the same
mAP--mIoU dissociation appears for OVAM. The MM-DiT captures do not face this
problem, and refining them yields the strongest or competitive results in every case. \methodname{} improves both base captures on all six benchmarks, by $\gain{+4.0}$ mIoU,
$\gain{+5.3}$ pAcc and $\gain{+7.0}$ mAP on average for ConceptAttention and $\gain{+4.5}$, $\gain{+6.1}$
and $\gain{+6.0}$ for Seg4Diff. We further notice that refinement matters most where raw attribution is weakest. On ADE20K
and Cityscapes the unrefined ConceptAttention capture falls behind TextSpan,
and it is refinement that returns the generative backbone to the front, by $\gain{+4.9}$
and $\gain{+6.3}$ mIoU over the best discriminative baseline. \methodname{} therefore sets
the state of the art among training-free OVSS methods, while being plug-and-play for DiTs.

\begin{figure}[ht]
    
  \centering
  \includegraphics[width=\linewidth]{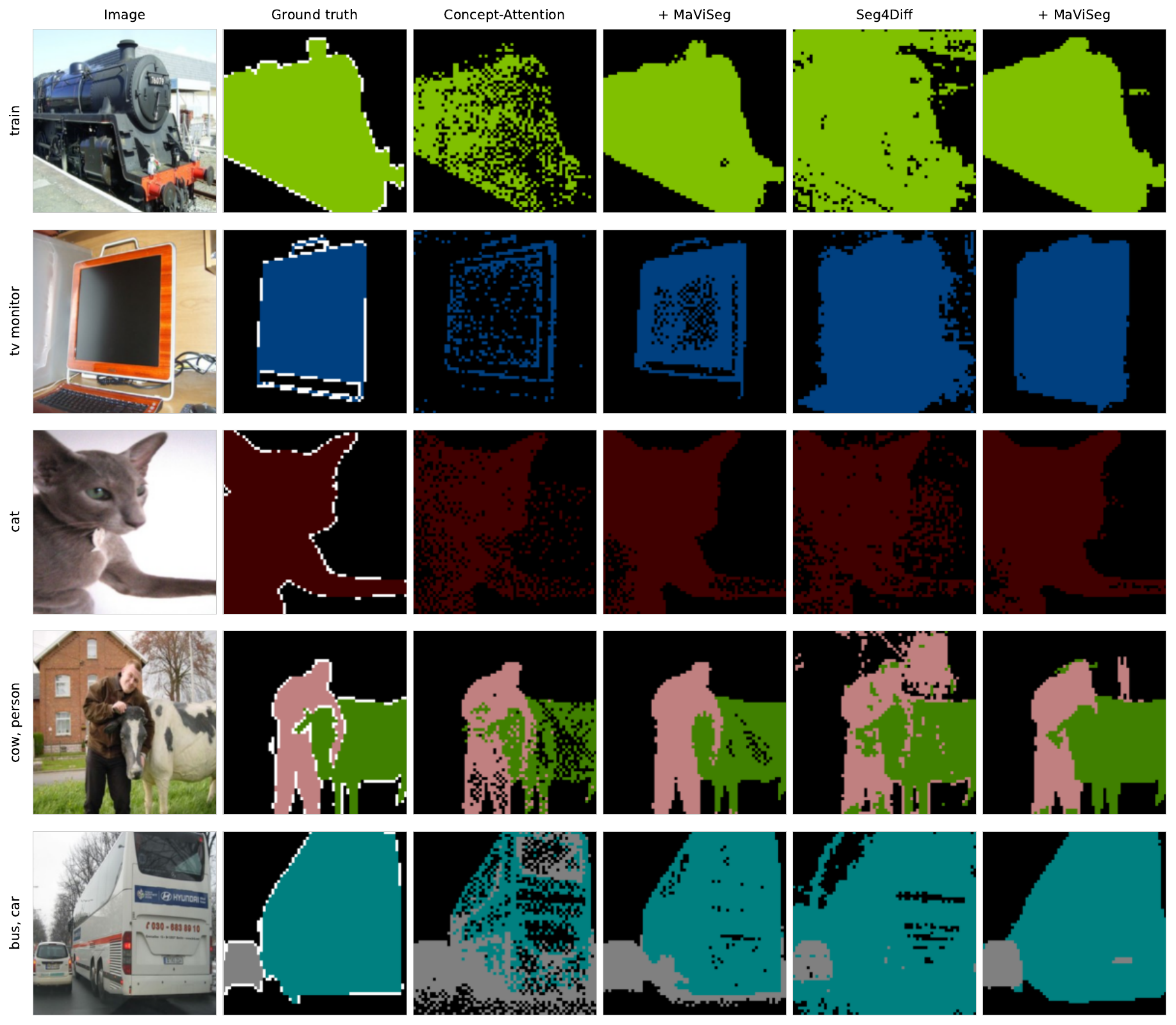}
  \caption{\textbf{Qualitative comparison of \methodname{} against existing methods.} \methodname{} fills hollow interiors and
  suppresses cross-edge bleed. Best viewed in zoom. Per-benchmark comparisons
  are in Appendix~\ref{supp:qualitative}.}
  \label{fig:qualitative}
\end{figure}

\paragraph{Qualitative examples.}
Fig.~\ref{fig:qualitative} contrasts the baseline capture with the
\methodname{}-refined read-out. The baseline shows its characteristic failures,
hollow object interiors, concept bleeding across edges,
while the refined maps fill interiors and sharpen boundaries toward the ground
truth. Comparisons on all six benchmarks, under both captures, are given in
Appendix~\ref{supp:qualitative}.

\begin{figure*}[t]
  \centering
  \includegraphics[width=\linewidth]{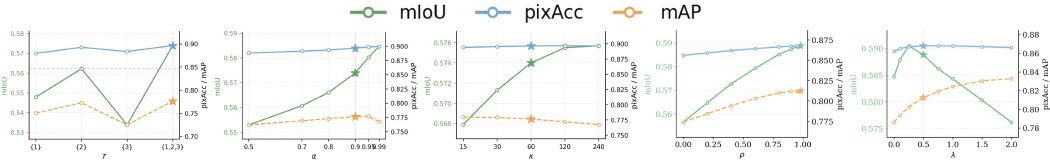}
  \caption{\textbf{Hyperparameter sweeps} on the full VOC validation set, one
  parameter varied with the others held at their final values. Stars mark the
  setting used throughout. Sweeps for $m$ and $|\mathcal{U}|$, and per-metric
  tables for every axis, are in Appendix~\ref{supp:hparams}.}
  \label{fig:hparams}
\end{figure*}

\subsection{Ablation Studies}

\input{Tables/components_ablation}
\paragraph{Per-component breakdown.}
We quantify each component's contribution to the final \methodname{} architecture
in Table~\ref{tab:ablation}. Read alone against
the baseline (row~1), \diff{MD} (row~2) is the strongest single operator on the
decision metrics, \ens{TDE} (row~3) helps more modestly, and \prot{BVP} (row~4)
sits below the baseline on mIoU and pixel accuracy while still raising mAP, so its
value is in the ranking rather than in the hard decision. The operators are
complementary. Each pair (rows~5--7) improves on its constituents, with
\diff{MD}+\ens{TDE} (row~5) giving the best mIoU and pixel accuracy short of the
full model and \diff{MD}+\prot{BVP} (row~6) giving the best mAP short of it. Thus, \diff{MD} drives the decision while \prot{BVP}
restores the fine ranking that a purely diffusion-smoothed field leaves under-resolved.
Enabling all three (row~8) is best on every metric. Per-class IoU for the
capture and for \methodname{} is given in Appendix~\ref{supp:perclass}.

\paragraph{Hyperparameters.} Each operator contributes its own parameters:
the capture set $\mathcal{T}$ from \ens{TDE}; the harvest fraction $r$, pool
size $|\mathcal{U}|$ and text-channel blend $\lambda$ from \prot{BVP}; and the
propagation weight $\alpha$, neighbourhood size $\kappa$, bandwidth index $m$
and residual blend $\rho$ from \diff{MD}. We show 5 primary hyperparameters for \methodname{} in Fig.~\ref{fig:hparams} and each is flat or monotone over its useful range, so one setting transfers across captures without per-dataset retuning. The timestep ensemble $\tau$ in \ens{TDE} is the clearest case, as aggregating over the full set clears the best single timestep, since no individual noise level is reliable on its own. In \diff{MD}, the propagation coefficient $\alpha$ raises the decision metrics monotonically, but mAP turns over once propagation gets too aggressive, so we take a high value just below that point rather than the maximum. The neighborhood size $\kappa$ is a mild trade-off in the same spirit, since mIoU rises and saturates as the graph grows while mAP falls throughout, so we stop where mIoU has mostly arrived and mAP has not yet given much up. The harvest fraction $r$ is the one axis with an interior optimum, and all
three metrics locate it at the same value, which
Appendix~\ref{supp:bank} attributes to the loss of separability between
prototype rows as the harvest grows. All three optima are broad, so \methodname{} is insensitive to the exact settings, and we defer the full per-metric sweeps, the remaining parameters $m$ and $|\mathcal{U}|$, the effect of transferring a bank between benchmarks, and the cost per stage to
Appendices~\ref{supp:hparams},~\ref{supp:transfer} and~\ref{supp:cost}.

\begin{figure}[ht]
  \centering
  \includegraphics[width=\linewidth]{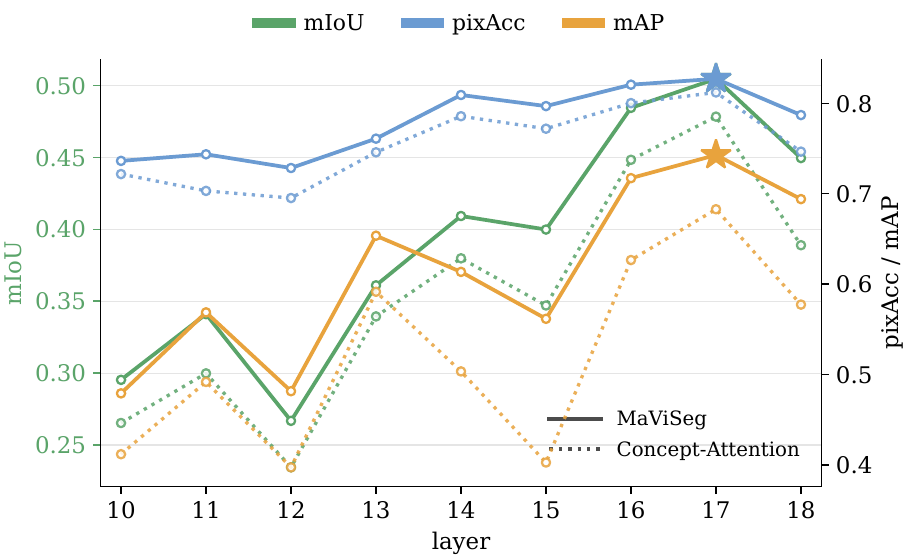}
  \caption{\textbf{Layer-wise improvement of \methodname{} over baselines.} We show trajectories of mIoU, pixel accuracy and mAP for deep layers 10-18 of the FLUX backbone ConceptAttention (CA). Results displayed are for the PASCAL-VOC val set. Notice how at each layer, the corresponding metric of CA is outperformed by \methodname{}.}
  \label{fig:layerwise}
\end{figure}
\paragraph{Layer-wise improvements.}Fig~\ref{fig:layerwise} repeats the read-out from each deep block in
isolation, with and without refinement. The base capture is uneven across the
range, alternating between sharp peaks and troughs from one block to the next,
and \methodname{} improves every block on all three metrics while preserving
that ordering, so it recovers signal already present in each layer rather than
substituting a preferred one. The improvement is widest at the weak blocks and
narrowest at the strongest one, most visibly in mAP, where the deepest troughs
close substantially. Refinement therefore flattens the layer dependence of the
read-out, and the reported result does not hinge on a hand-picked block.

%% file: Tables/main_results.tex
\begin{table*}[ht]
\centering
\setlength{\tabcolsep}{2.05pt}
\renewcommand{\arraystretch}{1.2}
\resizebox{\textwidth}{!}{%
\begin{tabular}{ll ccc ccc ccc ccc ccc ccc}
\toprule
 & & \multicolumn{3}{c}{VOC} & \multicolumn{3}{c}{PC-59} & \multicolumn{3}{c}{ADE20K} & \multicolumn{3}{c}{Cityscapes} & \multicolumn{3}{c}{COCO-Object} & \multicolumn{3}{c}{COCO-Stuff-27} \\
\cmidrule(lr){3-5} \cmidrule(lr){6-8} \cmidrule(lr){9-11} \cmidrule(lr){12-14} \cmidrule(lr){15-17} \cmidrule(lr){18-20}
Method & Back. & mIoU & pAcc & mAP & mIoU & pAcc & mAP & mIoU & pAcc & mAP & mIoU & pAcc & mAP & mIoU & pAcc & mAP & mIoU & pAcc & mAP \\
\midrule
\multicolumn{20}{@{}l}{\textbf{\textit{Discriminative}}} \\[1pt]
\cmidrule(l){1-20}
TransInterp & ViT & 0.192 & 0.677 & 0.476 & 0.258 & 0.455 & 0.373 & 0.207 & 0.394 & 0.259 & 0.159 & 0.354 & 0.163 & 0.171 & 0.626 & 0.454 & 0.177 & 0.342 & 0.269 \\
GradCAM & ViT & 0.397 & 0.737 & 0.503 & 0.405 & 0.624 & 0.584 & 0.303 & 0.538 & 0.519 & 0.227 & 0.515 & 0.392 & 0.290 & 0.659 & 0.584 & 0.266 & 0.449 & 0.378 \\
TextSpan & ViT & 0.434 & 0.727 & 0.727 & 0.421 & 0.641 & 0.619 & 0.324 & 0.559 & 0.518 & 0.235 & 0.507 & 0.301 & 0.308 & 0.647 & 0.624 & 0.270 & 0.453 & 0.435 \\
\addlinespace[3pt]
\cmidrule(l){1-20}
\multicolumn{20}{@{}l}{\textbf{\textit{Generative}}} \\[1pt]
\cmidrule(l){1-20}
DAAM & SDXL & 0.059 & 0.639 & 0.757 & 0.128 & 0.301 & 0.585 & 0.072 & 0.137 & 0.407 & 0.015 & 0.070 & 0.366 & 0.084 & 0.444 & 0.560 & 0.068 & 0.200 & 0.405 \\
OVAM & SD1.5 & 0.248 & 0.579 & 0.665 & 0.182 & 0.341 & 0.407 & 0.109 & 0.204 & 0.237 & 0.038 & 0.080 & 0.159 & 0.213 & 0.489 & 0.491 & 0.117 & 0.255 & 0.285 \\
DiffSegmenter & SD1.5 & 0.334 & 0.771 & \best{0.866} & 0.405 & 0.660 & 0.614 & 0.246 & 0.581 & 0.500 & 0.196 & \best{0.553} & 0.337 & 0.248 & 0.762 & 0.646 & 0.260 & 0.460 & 0.413 \\
CA & FLUX & 0.527 & \second{0.877} & 0.711 & 0.446 & 0.691 & 0.514 & 0.222 & 0.514 & 0.341 & 0.190 & 0.410 & 0.257 & 0.313 & \second{0.774} & 0.569 & 0.308 & 0.522 & 0.374 \\
\textbf{CA + \methodname{}} & FLUX & \best{0.574} & \best{0.896} & 0.777 & \second{0.506} & \second{0.761} & 0.620 & 0.246 & \second{0.604} & 0.436 & 0.231 & 0.466 & 0.291 & \second{0.332} & \best{0.786} & 0.613 & \best{0.357} & \best{0.591} & 0.447 \\
S4D & SD3-m & 0.467 & 0.737 & 0.818 & 0.484 & 0.705 & \second{0.661} & \second{0.350} & 0.597 & \second{0.539} & \second{0.260} & 0.515 & \second{0.447} & 0.311 & 0.604 & \second{0.671} & 0.294 & 0.519 & \second{0.458} \\
\textbf{S4D + \methodname{}} & SD3-m & \second{0.532} & 0.797 & \second{0.863} & \best{0.538} & \best{0.766} & \best{0.729} & \best{0.373} & \best{0.664} & \best{0.606} & \best{0.298} & \second{0.541} & \best{0.517} & \best{0.356} & 0.691 & \best{0.730} & \second{0.338} & \second{0.583} & \best{0.510} \\
\bottomrule
\end{tabular}}
\caption{\textbf{OVSS performance across datasets.} We report mIoU, pAcc and mAP for two types of backbone (Back.) based methods: discriminative and generative. For discriminative methods, VIT backbone refers to ViT-H-14. CA refers to Concept-Attention, S4D refers to Seg4Diff. \colorbox{bestbg}{\textbf{Best}} and
\colorbox{secondbg}{\underline{second best}} are highlighted per column. \methodname{} achieves state-of-the-art performance generally across all benchmarks.}
\label{tab:main}
\end{table*}

%% file: Tables/components_ablation.tex
\begin{table}[t]
\small
\centering
\setlength{\tabcolsep}{5.5pt}
\begin{tabular}{@{}c ccc ccc@{}}
\toprule
\multicolumn{4}{c}{} & \multicolumn{3}{c}{CA (FLUX)} \\
\cmidrule(lr){5-7}
row \# & \diff{MD} & \ens{TDE} & \prot{BVP} & mIoU & pAcc & mAP \\
\midrule
1 & \xmark & \xmark & \xmark & 0.527 & 0.877 & 0.711 \\
2 & \cmark & \xmark & \xmark & 0.564 & 0.893 & 0.755 \\
3 & \xmark & \cmark & \xmark & 0.544 & 0.883 & 0.721 \\
4 & \xmark & \xmark & \cmark & 0.518 & 0.874 & 0.735 \\
\midrule
5 & \cmark & \cmark & \xmark & \second{0.570} & \second{0.894} & 0.759 \\
6 & \cmark & \xmark & \cmark & 0.562 & 0.893 & \second{0.773} \\
7 & \xmark & \cmark & \cmark & 0.539 & 0.882 & 0.745 \\
\midrule
8 & \cmark & \cmark & \cmark & \best{0.574} & \best{0.896} & \best{0.777} \\
\bottomrule
\end{tabular}
\caption{\textbf{\methodname{} operator ablation.}The capture is from Concept-Attention (FLUX-schnell),
on PASCAL VOC val set, (\colorbox{bestbg}{\textbf{best}} and
\colorbox{secondbg}{\underline{second}} per column). 
}
\label{tab:ablation}
\end{table}

%% file: Sections/6_Limitations.tex
\section{Limitations}
\label{sec:limitations}

Although \methodname{} is a training-free OVSS method, it has a few limitations.
The \prot{Bootstrapped Visual Prototypes} channel trades some mIoU for increased pixel ranking (mAP), and its bank needs to be harvested once from an
unlabelled pool. Refinement with \diff{MD} also builds one affinity graph per image, which is quadratic
in the token count. Finally, we demonstrate capture-agnosticity across multiple
DiT backbones, and the same interface extends to non-DiT generative backbones,
for which we give an initial discussion in Appendix \ref{supp:agnosticity}.

%% file: Sections/7_Conclusion.tex
\section{Conclusion}
\label{sec:conclusion}
We introduce \methodname{}, a training-free refinement framework for open-vocabulary
segmentation on frozen diffusion transformers. Across all six benchmarks it
improves both base captures on every metric and sets the state of the art among
training-free methods. \methodname{} offers a shift from treating
attribution as a search for a cleaner capture to treating it as a decoding
problem, since existing read-outs and DiT backbones already hold the concept
information but leave much of it dormant on the way to the mask. More broadly, we
expect this view to point toward generative models whose internal concept
representations are faithful enough to be read, edited, and trusted.

%% file: Sections/X_supplementary.tex
\newenvironment{inplace}
  {\par\medskip\noindent\minipage{\columnwidth}}
  {\endminipage\par\medskip}

\setcounter{topnumber}{2}
\setcounter{bottomnumber}{1}
\setcounter{totalnumber}{3}
\setcounter{dbltopnumber}{2}
\renewcommand{\topfraction}{0.9}
\renewcommand{\bottomfraction}{0.6}
\renewcommand{\textfraction}{0.07}
\renewcommand{\floatpagefraction}{0.6}
\renewcommand{\dbltopfraction}{0.9}
\renewcommand{\dblfloatpagefraction}{0.5}

\newpage

\appendix
\section*{Supplementary Material}

\section{Dataset Details}
\label{supp:datasets}
Table~\ref{supp:tab-datasets} summarises the six benchmarks we use to evaluate existing baselines and \methodname{}. We evaluate
exclusively on official validation splits and use no training or test-server
data at any stage. In our
implementation, background is scored on VOC \citep{everingham2015voc} and
COCO-Object \citep{lin2014coco,caesar2018cocostuff} but not on the remaining
four, so pixel accuracy on those two is dominated by a single residual class
and is comparable between them rather than against the background-free
benchmarks; background is also the one class for which an appearance anchor is
ill-defined, and we exclude it from the prototype bank $\Pi$. Further, images
are resized to the backbone's $1024\times1024$ input and read out on the
$64\times64$ token grid, so a token covers roughly $8\times6$ native pixels on
VOC and Pascal-Context \citep{mottaghi2014pascalcontext} but $32\times16$ on
Cityscapes \citep{cordts2016cityscapes}, whose frames are the largest and the
only ones with a $2\!:\!1$ aspect ratio; thin Cityscapes categories are
consequently at or below one token in width before any attribution is computed,
which bounds every method reading from this grid, the base captures and
\methodname{} alike. Finally, the six benchmarks are six vocabulary
conditions over fewer image distributions, since COCO-Object and COCO-Stuff-27
\citep{caesar2018cocostuff} are the same $5{,}000$ images under different
vocabularies and VOC and Pascal-Context-59 both draw on the PASCAL pool, and
they differ tenfold in size, so the $500$-image Cityscapes split carries the
widest uncertainty of the six.

\begin{inplace}
{\centering
\small
\begin{tabular}{lccc}
\toprule
Dataset & Images & Classes & Native res. \\
\midrule
PASCAL VOC 2012   & 1{,}449 & 20\,(+bg) & 500$\times$375$^{\dagger}$ \\
Pascal-Context-59 & 5{,}104 & 59        & 500$\times$375$^{\dagger}$ \\
ADE20K            & 2{,}000 & 150       & $\sim$1.3\,MP$^{\dagger}$ \\
Cityscapes        & 500     & 19        & 2048$\times$1024 \\
COCO-Object       & 5{,}000 & 80\,(+bg) & 640$\times$480$^{\dagger}$ \\
COCO-Stuff-27     & 5{,}000 & 27        & 640$\times$480$^{\dagger}$ \\
\bottomrule
\end{tabular}
\par}
\captionof{table}{\textbf{Dataset Statistics.} Reported numbers are for the official validation split which we evaluate on. (+bg) marks vocabularies scoring an explicit background class, so VOC and COCO-Object are read out over 21 and 81 labels. $^{\dagger}$ denotes variable image size, for which we give the typical value. MP for ADE20K refers for average Megapixels for the images.}
\label{supp:tab-datasets}
\end{inplace}

\section{Qualitative Results by Dataset}
\label{supp:qualitative}
\begin{figure*}[p]
\centering
\includegraphics[width=\textwidth]{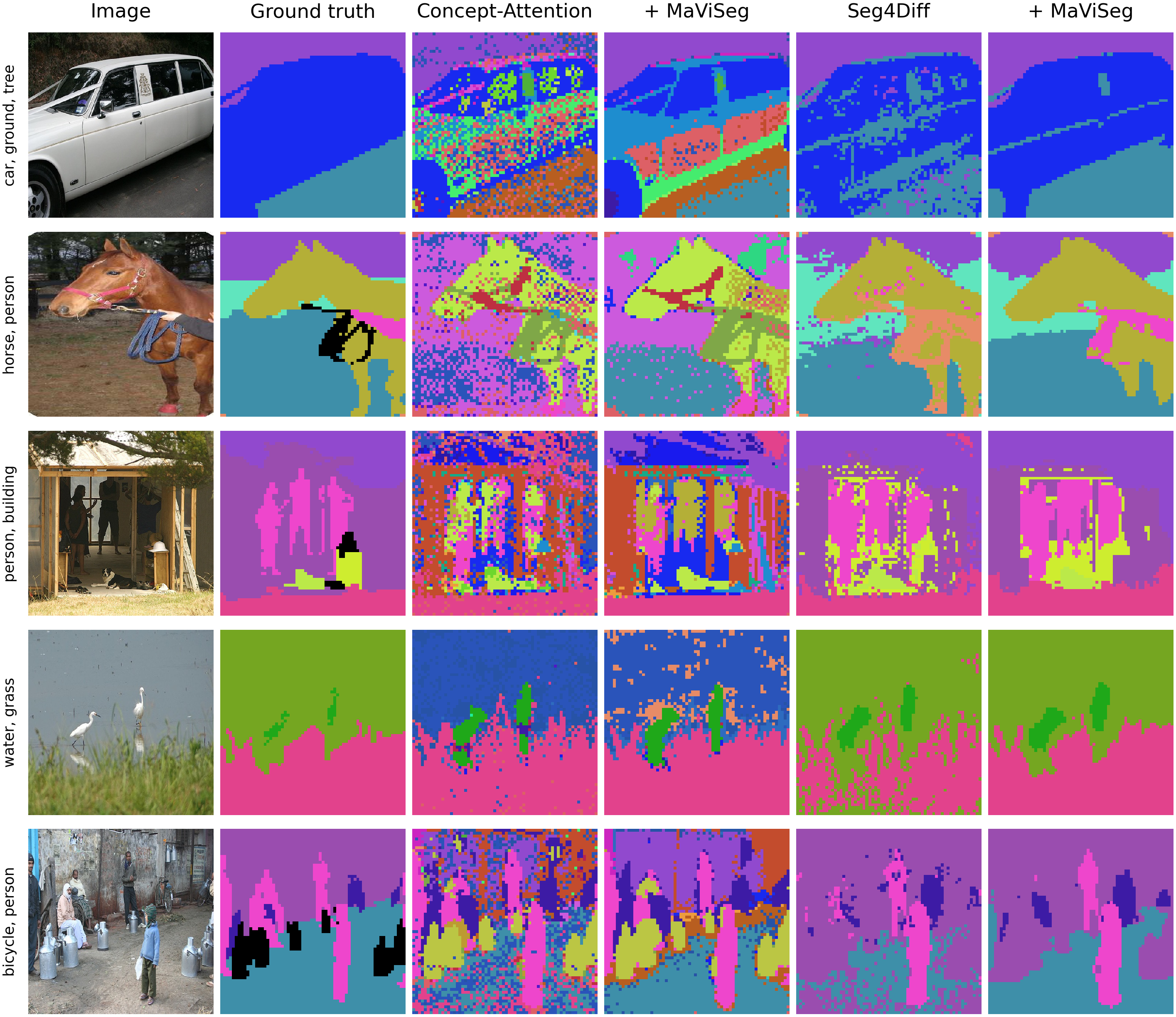}
\caption{\textbf{Qualitative results on Pascal-Context-59.} For each image (row), we show : Ground Truth, captures (by Concept-Attention and Seg4Diff) along with their improvements using \methodname{}. On the left of each image, we show the concepts (text) used to obtain semantic segmentation maps. Best viewed in color and in zoom.}
\label{supp:qual-pc59}
\end{figure*}

\begin{figure*}[p]
\centering
\includegraphics[width=\textwidth]{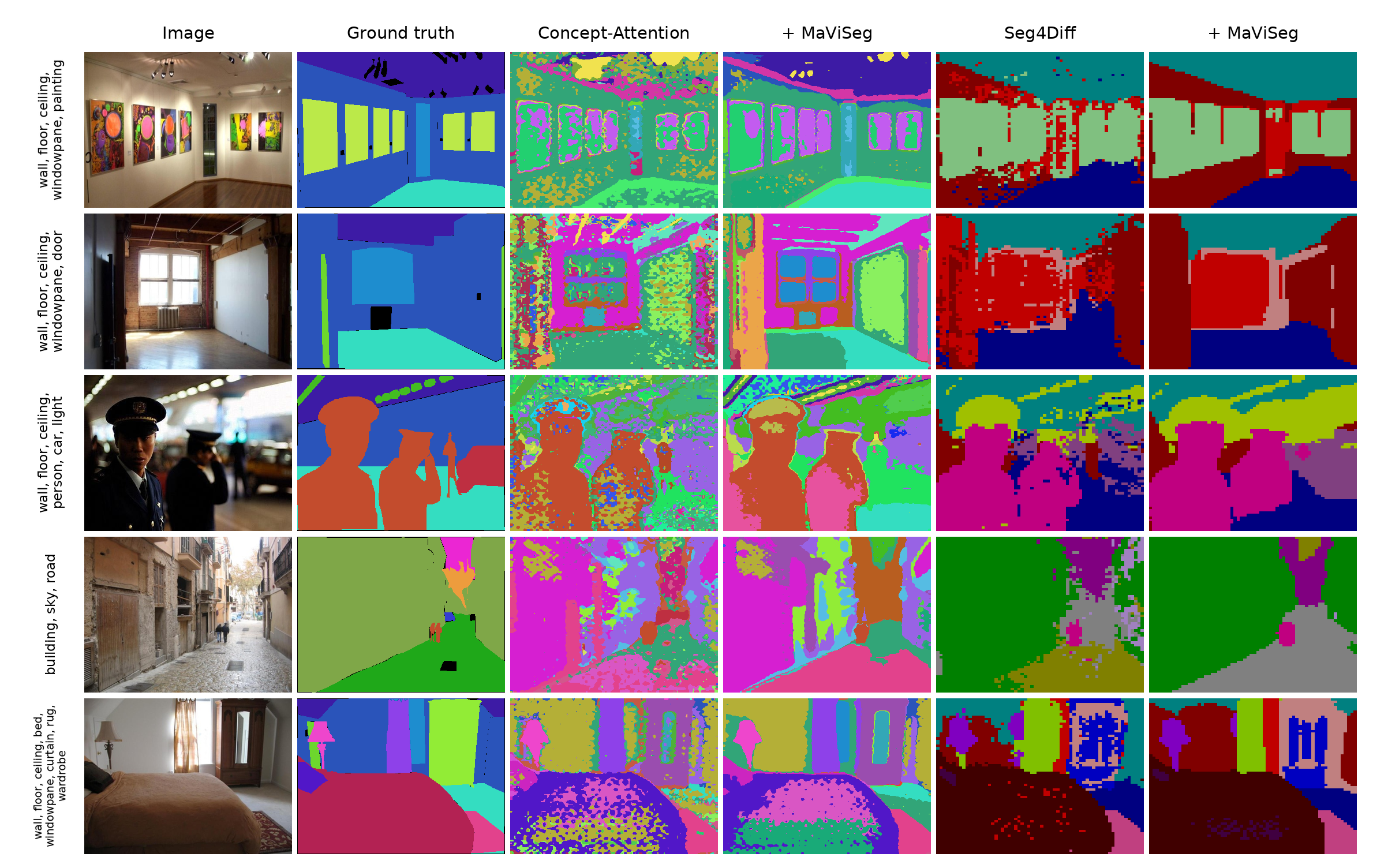}
\caption{\textbf{Qualitative results on ADE20K.} For each image (row), we show (columns) : Ground Truth, captures (by Concept-Attention and Seg4Diff) along with their improvements using \methodname{}. On the left of each image, we show the concepts (text) used to obtain semantic segmentation maps.}
\label{supp:qual-ade20k}
\end{figure*}

\begin{figure*}[p]
\centering
\includegraphics[width=\linewidth]{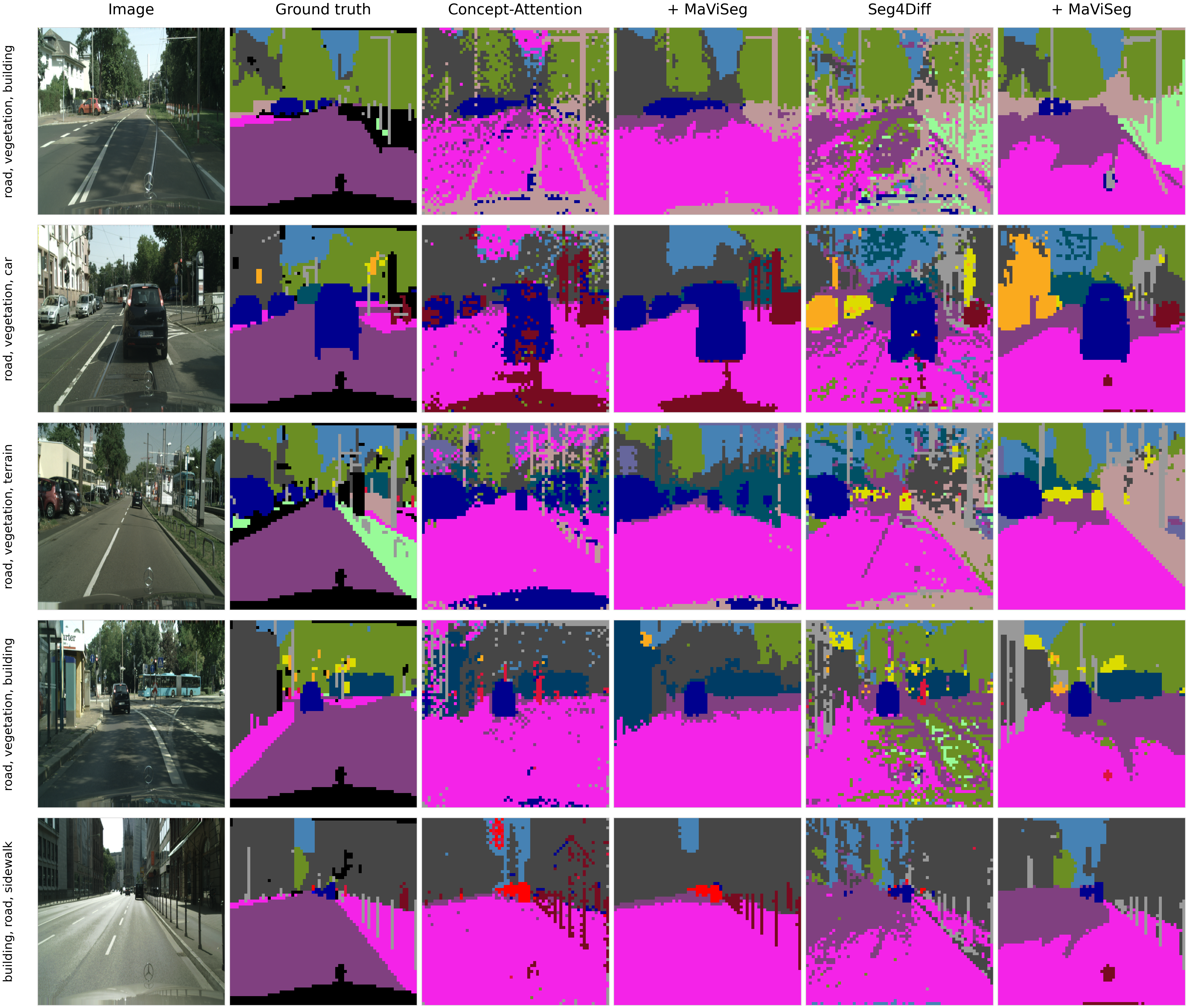}
\caption{\textbf{Qualitative results on Cityscapes.} For each image (row), we show (columns) : Ground Truth, captures (by Concept-Attention and Seg4Diff) along with their improvements using \methodname{}. On the left of each image, we show the concepts (text) used to obtain semantic segmentation maps. Best viewed in color and in zoom.}
\label{supp:qual-cityscapes}
\end{figure*}

\begin{figure*}[p]
\centering
\includegraphics[width=\textwidth]{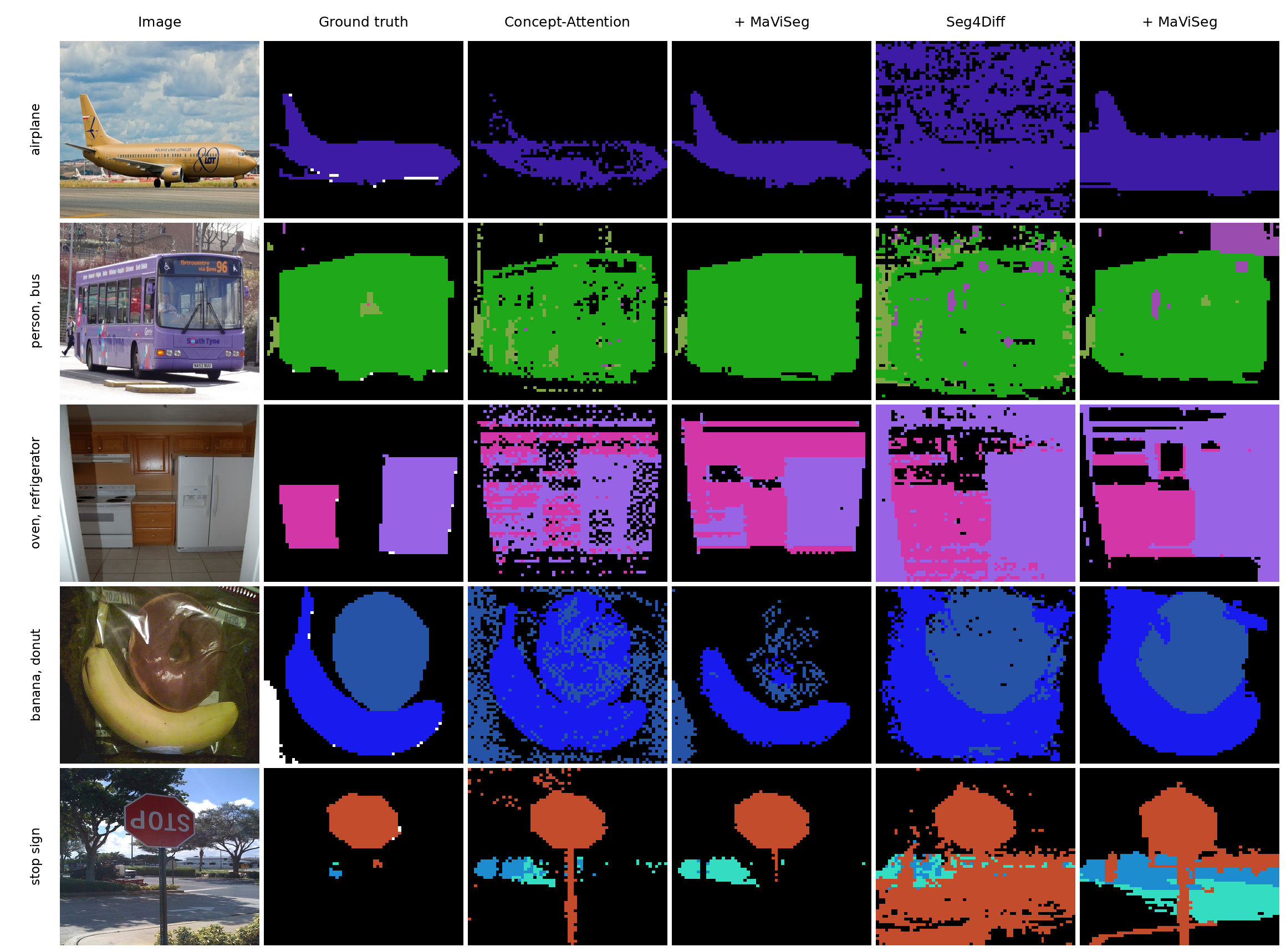}
\caption{\textbf{Qualitative results on COCO-Object.} For each image (row), we show (columns) : Ground Truth, captures (by Concept-Attention and Seg4Diff) along with their improvements using \methodname{}. On the left of each image, we show the concepts (text) used to obtain semantic segmentation maps. Best viewed in color and in zoom.}
\label{supp:qual-coco-object}
\end{figure*}

\begin{figure*}[p]
\centering
\includegraphics[width=\textwidth]{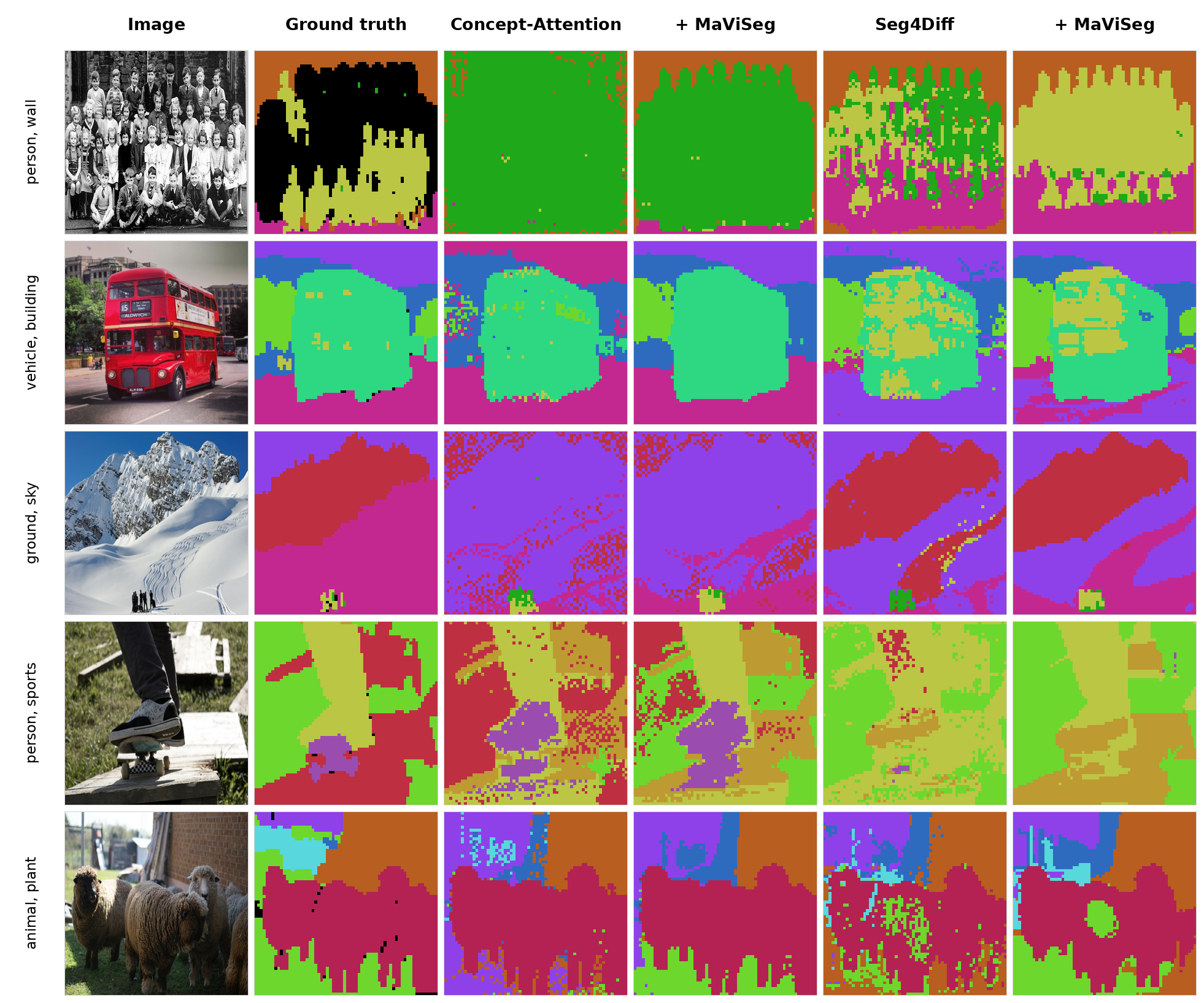}
\caption{\textbf{Qualitative results on COCO-Stuff-27.} For each image (row), we show (columns) : Ground Truth, captures (by Concept-Attention and Seg4Diff) along with their improvements using \methodname{}. On the left of each image, we show the concepts (text) used to obtain semantic segmentation maps. Best viewed in color and in zoom.}
\label{supp:qual-coco-stuff}
\end{figure*}

Figures~\ref{supp:qual-pc59}, \ref{supp:qual-ade20k},
\ref{supp:qual-cityscapes}, \ref{supp:qual-coco-object} and
\ref{supp:qual-coco-stuff} show validation examples for Pascal-Context-59,
ADE20K, Cityscapes, COCO-Object and COCO-Stuff-27 respectively, each under both
captures. The capture columns exhibit the characteristic failures, hollow
interiors and concept bleed across edges, which the paired \methodname{} columns
fill and sharpen toward the ground truth. That the same refinement improves both
captures, which read different internal signals, is the capture-agnosticity of
\S\ref{supp:agnosticity} shown rather than tabulated.


\section{Full Hyperparameter Sweeps}
\label{supp:hparams}

\begin{inplace}
{\centering
\small
\setlength{\tabcolsep}{5pt}
\begin{tabular}{@{}ll rrr@{}}
\toprule
Param. & Value & mIoU & pAcc & mAP \\
\midrule
$\mathcal{T}$
 & $\{1\}$        & 0.5478 & 0.8797 & 0.7504 \\
 & $\{2\}$        & 0.5622 & 0.8926 & 0.7726 \\
 & $\{3\}$        & 0.5338 & 0.8836 & 0.7249 \\
 & $\{1,2,3\}^{\dagger}$ & \textbf{0.5739} & \textbf{0.8965} & \textbf{0.7765} \\
\midrule
$\alpha$
 & 0.5            & 0.5529 & 0.8881 & 0.7618 \\
 & 0.7            & 0.5607 & 0.8913 & 0.7692 \\
 & 0.8            & 0.5660 & 0.8934 & 0.7728 \\
 & $0.9^{\dagger}$ & \textbf{0.5739} & \textbf{0.8965} & \textbf{0.7765} \\
\midrule
$\kappa$
 & 15             & 0.5679 & 0.8941 & \textbf{0.7796} \\
 & 30             & 0.5713 & 0.8955 & 0.7792 \\
 & $60^{\dagger}$ & 0.5740 & 0.8965 & 0.7765 \\
 & 120            & 0.5755 & \textbf{0.8969} & 0.7720 \\
 & 240            & \textbf{0.5757} & 0.8966 & 0.7674 \\
\midrule
$m$
 & 3              & 0.5715 & 0.8956 & 0.7734 \\
 & 5              & 0.5730 & 0.8961 & 0.7748 \\
 & $7^{\dagger}$  & 0.5740 & 0.8965 & 0.7765 \\
 & 10             & 0.5748 & 0.8968 & 0.7782 \\
 & 15             & \textbf{0.5757} & \textbf{0.8972} & \textbf{0.7798} \\
\midrule
$|\mathcal{U}|$
 & 150            & 0.5725 & 0.8962 & 0.7759 \\
 & $300^{\dagger}$ & 0.5740 & 0.8965 & 0.7765 \\
 & 450            & 0.5752 & 0.8974 & \textbf{0.7770} \\
 & 600            & \textbf{0.5768} & \textbf{0.8980} & 0.7765 \\
\midrule
$r$
 & $0.5\%$        & 0.5646 & 0.8947 & 0.7741 \\
 & $1\%$          & 0.5669 & 0.8951 & 0.7749 \\
 & $2\%^{\dagger}$ & \textbf{0.5740} & \textbf{0.8965} & \textbf{0.7765} \\
 & $5\%$          & 0.5719 & 0.8948 & 0.7748 \\
 & $10\%$         & 0.5695 & 0.8929 & 0.7724 \\
 & $20\%$         & 0.5633 & 0.8887 & 0.7681 \\
\midrule
$\rho$
 & 0              & 0.5390 & 0.8822 & 0.7451 \\
 & 0.2            & 0.5477 & 0.8859 & 0.7536 \\
 & 0.4            & 0.5560 & 0.8892 & 0.7620 \\
 & 0.6            & 0.5634 & 0.8923 & 0.7693 \\
 & 0.8            & 0.5695 & 0.8947 & 0.7743 \\
 & 0.9            & 0.5722 & 0.8958 & 0.7758 \\
 & 0.95           & 0.5734 & 0.8963 & 0.7763 \\
 & $0.98^{\dagger}$ & \textbf{0.5740} & \textbf{0.8965} & \textbf{0.7765} \\
\midrule
$\lambda$
 & 0              & 0.5695 & 0.8943 & 0.7589 \\
 & 0.1            & \textbf{0.5740} & \textbf{0.8966} & 0.7638 \\
 & $0.5^{\dagger}$ & \textbf{0.5740} & 0.8965 & 0.7765 \\
 & 0.75           & 0.5707 & 0.8950 & 0.7816 \\
 & 1.0            & 0.5673 & 0.8937 & 0.7851 \\
 & 1.5            & 0.5619 & 0.8916 & 0.7889 \\
 & 2.0            & 0.5579 & 0.8900 & \textbf{0.7910} \\
\bottomrule
\end{tabular}
\par}
\captionof{table}{Per-metric sweeps on the full VOC validation set ($1449$ images),
one parameter varied with the others held at their final values. $\dagger$
marks the value used throughout, and bold marks the best observed on each
metric within an axis.}
\label{supp:tab-hparams}
\end{inplace}

Table~\ref{supp:tab-hparams} gives all three metrics for every hyperparameter,
on the full PASCAL VOC validation set. Each sweep varies one parameter with the
others held at the values of \S4.1, so a row differs from the reference setting
in one place only. The main paper shows mIoU for $\mathcal{T}$, $\alpha$ and
$\kappa$ in Figure~4 and defers the rest here.
 
Performance is monotone increasing in
$\alpha$, $\kappa$, $m$, $|\mathcal{U}|$ and $\rho$ across the values shown, and
the response is shallow near the top of each: the last step in $\rho$, from
$0.95$ to $0.98$, is worth $+0.0006$ mIoU against $+0.0087$ for the first step
from $0$ to $0.2$. The graph parameters behave the same way, with $\kappa$
spanning $0.8$ mIoU over a sixteen-fold range and $m$ only $0.4$ over a
five-fold one, so neither needs per-dataset tuning. The exception is the pair that controls the prototype channel. The harvest
fraction $r$ has an interior optimum at $2\%$, where all three metrics peak
together, and the blend $\lambda$ separates the metrics outright.
mAP increases monotonically across the whole range, by $+0.032$ from
$\lambda=0$ to $\lambda=2$, while mIoU is flat between $0.1$ and $0.5$ and then
falls by $0.016$ over the rest. Weighting the prototype channel more heavily
keeps improving how the score field ranks concepts and keeps degrading the
arg-max taken from it. This is the operator-level form of what
Table~\ref{tab:ablation} shows when \prot{BVP} is enabled in isolation, and it
is the clearest instance of that trade in the paper. We use $\lambda=0.5$,
which sits at the top of the mIoU plateau and recovers a substantial part of
the mAP range.

\section{Convergence of the Diffusion Iteration}
\label{supp:convergence}
 
\begin{figure}[ht]
{\centering
\includegraphics[width=\linewidth]{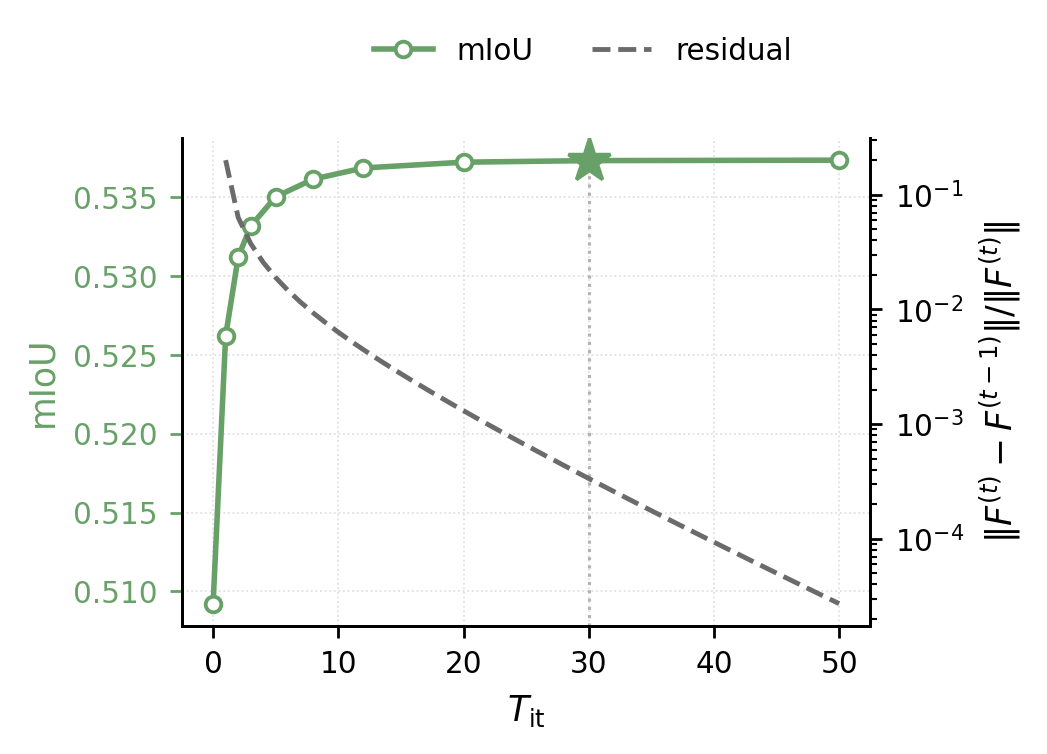}
\par}
\captionof{figure}{The residual falls geometrically and mIoU is indistinguishable from
its limit well before the setting used. The star marks
$T_{\mathrm{it}}=30$.}
\label{supp:fig-conv}
\end{figure}

$T_{\mathrm{it}}$ is fixed rather than swept, on the grounds that the power
iteration of Eq.~(9) converges. Table~\ref{supp:tab-conv} and
Fig.~\ref{supp:fig-conv} measure that two ways: mIoU as a function of the
iteration count, and the relative residual
$\lVert F^{(t)} - F^{(t-1)}\rVert_F / \lVert F^{(t)}\rVert_F$, which states the
same thing numerically without reference to any metric.
 
\begin{table}[ht]
{\centering
\small
\begin{tabular}{@{}rrrr@{}}
\toprule
$T_{\mathrm{it}}$ & mIoU & pAcc & residual \\
\midrule
0  & 0.5092 & 0.8659 & n/a \\
1  & 0.5262 & 0.8738 & $2.0\times10^{-1}$ \\
2  & 0.5312 & 0.8760 & $6.4\times10^{-2}$ \\
3  & 0.5332 & 0.8769 & $3.8\times10^{-2}$ \\
5  & 0.5350 & 0.8777 & $1.9\times10^{-2}$ \\
8  & 0.5361 & 0.8781 & $9.3\times10^{-3}$ \\
12 & 0.5368 & 0.8783 & $4.5\times10^{-3}$ \\
20 & 0.5372 & 0.8784 & $1.3\times10^{-3}$ \\
$30^{\dagger}$ & 0.5373 & 0.8785 & $3.4\times10^{-4}$ \\
50 & 0.5373 & 0.8785 & $2.7\times10^{-5}$ \\
\bottomrule
\end{tabular}
\par}
\captionof{table}{Quality and residual against iteration count, on a $300$ image subset
of the VOC validation set with every other hyperparameter at its final value.
$\dagger$ marks the setting used. Absolute mIoU is lower than
Table~\ref{tab:main} because the subset is smaller, and the claim here concerns
the shape of the curve rather than its level. $T_{\mathrm{it}}=0$ is the
undiffused field $\tilde{\mathbf{S}}$.}
\label{supp:tab-conv}
\end{table}
 
Most of what diffusion contributes arrives immediately. The first iteration is
worth $+1.7$ mIoU points and the first three account for $85\%$ of the total
$+2.8$, after which the curve flattens. mIoU comes within $0.0005$ of its final
value at $T_{\mathrm{it}}=12$ and does not move at all after $30$, so the
setting we use sits $18$ iterations past the point where further iteration
stops changing the read-out.
 
The residual makes the same point without reference to a metric, and its rate
is the one the iteration predicts of itself. Writing
$\mathbf{e}^{(t)} = \mathbf{F}^{(t)} - \mathbf{F}^{\star}$, Eq.~(9) gives
$\mathbf{e}^{(t+1)} = \alpha \mathbf{P}\mathbf{e}^{(t)}$, and since
$\mathbf{P}$ is similar to a random-walk matrix its spectral radius is at most
$1$, so the error contracts by at most $\alpha$ per step. The measured
asymptotic ratio is $0.882$ against the bound of $\alpha=0.9$, which is what a
correct implementation should show. Early steps contract faster, at $0.32$ for
the first, because the initial error has substantial weight on the lower part
of the spectrum.
 
Fixing $T_{\mathrm{it}}$ generously rather than tuning it costs almost nothing.
Diffusion is $3.6$\,ms per image at $30$ iterations
(Appendix~\ref{supp:cost}), so a single iteration is $0.12$\,ms and the $18$
past the point of convergence add $2.2$\,ms, under $0.1\%$ of the pipeline. We
therefore fix $T_{\mathrm{it}}=30$ rather than treating it as a hyperparameter,
and it is the one setting in the method that could be raised without
consequence.

\section{Prototype Bank Analysis}
\label{supp:bank}
 
Table~\ref{supp:tab-bank} measures how separable the prototype rows are as the
harvest fraction $r$ grows. Two statistics summarise a bank
$\bm{\Pi}\in\mathbb{R}^{K\times d}$ whose rows have been $\ell_2$-normalised.
The mean off-diagonal cosine is the average similarity between distinct
prototypes, so $0$ means they point in unrelated directions and $1$ means they
have collapsed onto one. The effective rank is the participation ratio of the
singular spectrum, $(\sum_i\sigma_i)^2/\sum_i\sigma_i^2$, which equals $K$ when
the bank spans every available direction and $1$ when it spans a single one.
 
\begin{table}[ht]
{\centering
\begin{tabular}{@{}lccc@{}}
\toprule
$r$ & mean cos. & eff.\ rank & mIoU \\
\midrule
$0.5\%$ & 0.5915 & 11.97 & 0.5646 \\
$1\%$   & 0.6006 & 11.63 & 0.5669 \\
$2\%$   & \textbf{0.6212} & \textbf{11.01} & \textbf{0.5740} \\
$5\%$   & 0.6517 & 10.13 & 0.5719 \\
$10\%$  & 0.6901 &  9.15 & 0.5695 \\
$20\%$  & 0.7513 &  7.74 & 0.5633 \\
\bottomrule
\end{tabular}
\par}
\captionof{table}{Prototype separability against the harvest fraction, on the VOC
vocabulary of $K=27$ concepts, with mIoU on the full validation set for
reference. Bold marks the setting used throughout.}
\label{supp:tab-bank}
\end{table}
 
\begin{figure}
{\centering
\includegraphics[width=\linewidth]{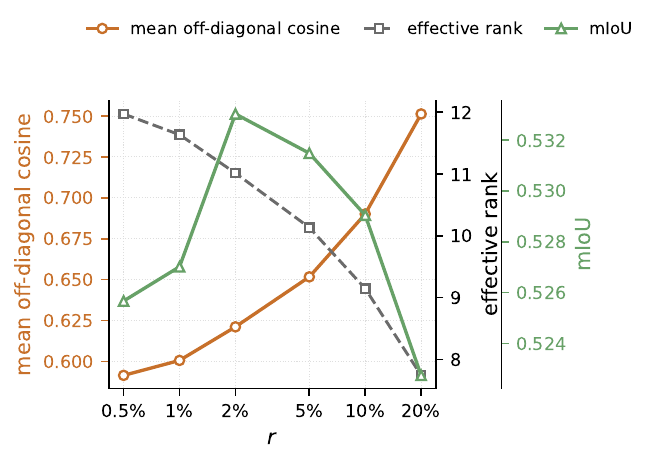}
\par}
\captionof{figure}{Separability falls monotonically as the harvest grows, while mIoU
peaks at $r=2\%$. The decline in mIoU past that point tracks the decline in
separability. The rise below it does not, and is attributable to the variance
of a prototype estimated from very few tokens rather than to how distinct the
prototypes are.}
\label{supp:fig-bank}
\end{figure}
 
Separability degrades monotonically with $r$. The mean off-diagonal cosine
rises from $0.5915$ to $0.7513$ and the effective rank falls from $11.97$ to
$7.74$ out of $27$, so a bank harvested at $20\%$ spans barely more than a
quarter of the directions available to it. This is the mechanism behind the
right-hand side of the optimum in Figure~4. A large harvest takes the top $r$
fraction of tokens for every concept in every pool image, and since the scorer
has no labels it cannot skip images in which a concept does not appear, so the
surplus tokens belong to no instance of the concept and pull every prototype
toward a common direction.
 
The left-hand side has a different cause. Separability is highest at
$r=0.5\%$, where mIoU is lowest, so the loss there is not one of confusion
between concepts. It is the variance of an anchor estimated from roughly
$20$ tokens per concept per image. The two effects act in opposite directions,
which is why $r$ is the only hyperparameter in Table~\ref{supp:tab-hparams} with an
interior optimum rather than a monotone or flat response.
 
\paragraph{Which prototypes collapse first.}
At every value of $r$ the closest pair is \emph{wall} and \emph{background},
rising from $0.959$ to $0.992$ cosine. Both are background words, and their
convergence is expected, since the two describe the same thing for most images
and the harvest has no way to separate them. The collapse is therefore
concentrated where it costs least.
 

\section{Prototype Bank Transfer}
\label{supp:transfer}

The prototype channel needs an unlabelled pool, and \S3.4 harvests it from
training images of the dataset under evaluation. Whether the pool has to come
from that dataset is a separate question from whether it has to be unlabelled,
and it decides how widely the method applies, since a practitioner with no
in-domain images could still use the text channel but not the appearance one.

We hold the vocabulary fixed and vary only the source of the pool. All three
rows of Table~\ref{supp:tab-transfer} use the same $27$ VOC concepts and are
evaluated on the full VOC validation set at the settings of \S4.1, so they
differ in one thing only. The in-domain bank is harvested from VOC training
images, disjoint from the evaluation split. The transferred bank is harvested
from ADE20K, chosen because it is scene-centric where VOC is object-centric, so
several VOC classes appear rarely or not at all in it. This is the harder
direction rather than a neutral one.

\begin{inplace}
{\centering
\small
\setlength{\tabcolsep}{4pt}
\resizebox{\columnwidth}{!}{%
\begin{tabular}{@{}l rrr rr@{}}
\toprule
Harvest pool & mIoU & pAcc & mAP & cos. & rank \\
\midrule
none                 & 0.5775 & 0.8971 & 0.7707 & -- & -- \\
VOC (in-domain)      & \textbf{0.5839} & \textbf{0.9006} & \textbf{0.7880}
                     & 0.621 & 11.0 \\
ADE20K (transferred) & 0.5813 & 0.8992 & 0.7811 & 0.634 & 10.6 \\
\bottomrule
\end{tabular}}
\par}
\captionof{table}{Prototype pool transfer on the full VOC validation set. All rows use
the VOC vocabulary and the settings of \S4.1 and differ only in which images
the bank was harvested from. \emph{cos.} is the mean off-diagonal cosine
between prototype rows and \emph{rank} the effective rank out of $K=27$, both
as defined in Appendix~\ref{supp:bank}. The three rows come from a single run
and are intended to be read against one another.}
\label{supp:tab-transfer}
\end{inplace}

The transferred bank recovers most of the benefit but not all of it. Against
the no-prototype row it retains $59\%$ of the in-domain gain on mIoU, $60\%$ on
pixel accuracy and $60\%$ on mAP. That the three agree to within a percentage
point is the more informative fact, because the mIoU differences here are small
in absolute terms and a single metric could carry that agreement by accident.
The pool therefore does not have to be drawn from the dataset under evaluation,
and it does have to resemble it.

The bank statistics say where the missing $40\%$ goes. The transferred bank is
slightly less separable than the in-domain one, at $0.634$ mean cosine against
$0.621$ and effective rank $10.6$ against $11.0$ out of $27$. Placed on the
harvest-fraction curve of Appendix~\ref{supp:bank}, where cosine runs from
$0.592$ to $0.751$, the transferred bank sits where an in-domain bank harvested
at roughly $r=3.3\%$ would, so the degradation is real but mild. This is the
mechanism the harvest predicts of itself. A concept with no instance in the
pool still receives a prototype, assembled from whichever tokens the text
channel scored highest, and those tokens belong to something else. The bank
does not lose the concepts ADE contains, it dilutes the ones it does not.

The appearance anchors are substantially a
property of the concept rather than of the dataset, which widens where the
method applies, since the pool is its only cross-image input. And the pool must
cover the vocabulary, which sharpens the sentence in \S5: a concept missing
from the pool falls back on the text channel, and a concept thinly represented
contributes a diluted anchor rather than none at all.

\section{Structure of the Affinity Graph}
\label{supp:affinity}
 \begin{table}
{\centering
\small
\begin{tabular}{@{}rrrrrr@{}}
\toprule
$\kappa$ & degree & in-class & conc. & mIoU & mAP \\
\midrule
15  &  18.7 & 0.762 & 10.4 & 0.5679 & \textbf{0.7796} \\
30  &  37.7 & 0.738 &  9.9 & 0.5713 & 0.7792 \\
$60^{\dagger}$ &  76.0 & 0.715 &  9.3 & 0.5740 & 0.7765 \\
120 & 153.0 & 0.692 &  8.9 & 0.5755 & 0.7720 \\
240 & 305.6 & 0.669 &  8.4 & \textbf{0.5757} & 0.7674 \\
\bottomrule
\end{tabular}
\par}
\captionof{table}{Locality of the feature graph against neighbourhood size, over $101$
VOC validation images and $222$ class instances. \emph{Degree} is the mean
number of non-zero entries per row, \emph{in-class} the share of a class's
affinity mass landing on tokens of the same class, and \emph{conc.} that share
divided by the share of tokens the class occupies. mIoU and mAP are the full
validation set values from Table~\ref{supp:tab-hparams}. $\dagger$ marks the
setting used.}
\label{supp:tab-affinity}
\end{table}
The manifold operator is built on the claim that the feature graph keeps
propagation inside the region an object occupies. Table~\ref{supp:tab-affinity}
measures that. For every class present in an image we compute the share of its
affinity mass that lands on tokens of the same class, and divide by the share of
tokens the class occupies. That ratio is the concentration: a value of $1$ means
the graph is indifferent to class, and higher values mean mass is retained
inside the object. Statistics are over $101$ validation images and $222$ class
instances, with all other hyperparameters at their final values.

The graph is strongly local at every setting tested, holding
between $8.4$ and $10.4$ times more affinity inside a class than its size alone
would give, so propagation is confined to the object whether the neighbourhood
is $15$ or $240$. And locality decreases monotonically as the neighbourhood
grows, from $10.4$ to $8.4$, a fall of $19\%$, with in-class mass dropping
$9.3$ points over the same range.
 
\begin{figure}
{\centering
\includegraphics[width=\linewidth]{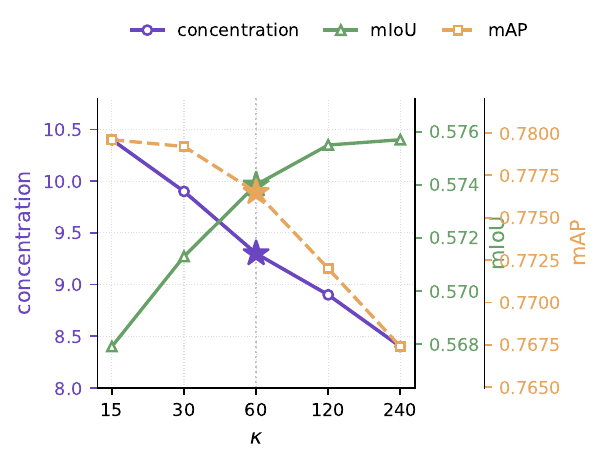}
\par}
\captionof{figure}{Locality and the metrics against neighbourhood size. Concentration and
mAP fall together as the graph grows while mIoU rises, so $\kappa$ selects a
point on a trade rather than a threshold. Stars mark the setting used.}
\label{supp:fig-affinity}
\end{figure}

Set against the metrics, this is what reconciles $\kappa$ with the rest of the
method. Figure~\ref{supp:fig-affinity} shows the three curves together. Across
the whole range, mAP moves with concentration and mIoU moves against it. A denser graph reaches further and recovers more of each object,
which the decision metrics reward, while admitting more cross-class edges, which
the ranking metric penalises. Sparsity is therefore not what makes the operator
work, since it works at every $\kappa$ we tried. Sparsity controls how tightly
the propagation is confined, and $\kappa$ selects a point on that trade rather
than a threshold below which the operator would fail. This is the same
ranking-against-decision tension that $\lambda$ shows in
Appendix~\ref{supp:hparams} and that \prot{BVP} shows in
Table~\ref{tab:ablation}, arriving here through a third mechanism. Symmetrising with
$\mathbf{W} \leftarrow \max(\mathbf{W}, \mathbf{W}^{\top})$ gives a token edges
from every other token that selected it, so the realised degree exceeds $\kappa$
by a factor of roughly $1.27$ and is stable across the range. The sparsity
figure quoted in \S3.5, $O(\kappa N)$ non-zeros, holds up to that constant.

\section{Per-Class Results}
\label{supp:perclass}

\begin{figure}[ht]
    {\centering
    \includegraphics[width=\linewidth]{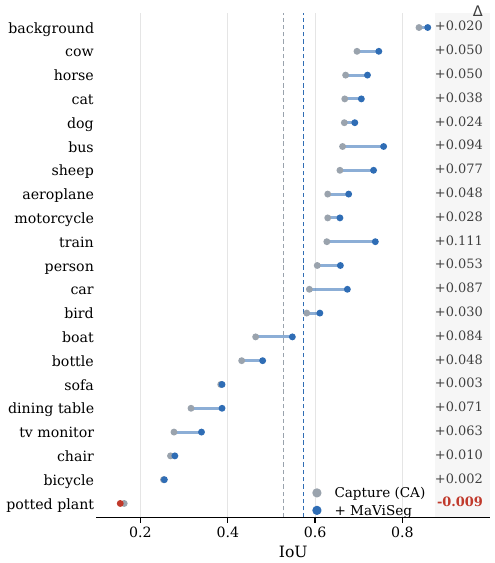}
\par}
    \captionof{figure}{\textbf{Per-class IoU on the full VOC validation set}, for the
    unrefined \textsc{ConceptAttention} capture (grey) and for \methodname{}
    (blue), from the run producing rows~1 and~8 of Table~\ref{tab:ablation}.
    Classes are ordered by the IoU of the capture, and each segment is the gain
    refinement adds. Dashed lines mark the two means. Twenty of twenty-one
    classes improve; the sole regression, \emph{potted plant}, is drawn in red
    and points the other way. The segments are visibly shortest at the foot of
    the ordering, among the thin and perforated categories that present little
    interior for \diff{manifold diffusion} to fill.}
    \label{fig:perclass}
\end{figure}

Refinement helps almost everywhere, improving $20$ of $21$ classes for a mean
gain of $4.7$ points, but Figure~\ref{fig:perclass} shows the gain is far from
uniform and the pattern is informative. The longest segments belong to
\emph{train} ($+0.111$), \emph{bus} ($+0.094$), \emph{car} ($+0.087$) and
\emph{boat} ($+0.084$); the shortest to \emph{bicycle} ($+0.002$),
\emph{sofa} ($+0.003$) and \emph{chair} ($+0.010$), and the only class to
regress is \emph{potted plant} ($-0.009$).

This is what the geometric operator predicts of itself. \diff{Manifold
diffusion} propagates concept mass through a token graph, and its
characteristic repair is a filled interior, so a class should benefit in
proportion to how much interior it has. Bicycles, chairs and potted plants are
thin, wiry or perforated: they present almost no interior to fill, and most of
their pixels lie on a boundary that propagation is designed not to cross.
Morphology separates the classes more sharply than difficulty does. Among the
seven classes the capture handles worst, the three thin ones gain $+0.1$ points
on average while the four with solid interiors gain $+4.6$, despite comparable
starting IoU. Difficulty alone is a poor predictor: the correlation between a
class's unrefined IoU and its gain is only $r=0.35$ ($95\%$ CI $[-0.10, 0.68]$,
$n=21$), and grouping by capture IoU into tertiles gives $+2.7$, $+6.3$ and
$+5.0$ points, so the weakest classes do gain least but the progression is not
monotone. What refinement leaves behind is therefore structural rather than
semantic, and it marks the limit of an operator that repairs extent rather than
identity.

\section{Computational Cost}
\label{supp:cost}
 
Table~\ref{supp:tab-cost} times each stage of Algorithm~1 separately on the
configuration used throughout, measured over $50$ VOC validation images on a
single NVIDIA A100. Each stage is timed with the CUDA stream synchronised, and
the first image is discarded as warm-up. The grid is $N=4096$ tokens with
$d=3072$, and $K$ varies per image with the concept set.

\begin{table}[ht]
{\centering
\small
\setlength{\tabcolsep}{4pt}
\begin{tabular}{@{}lrrr@{}}
\toprule
Stage & Time (ms) & \% & Mem. (GB) \\
\midrule
Capture ($|\mathcal{T}|{=}3$)  & 2578.1 & 99.5 & 2.9 \\
\midrule
\prot{BVP} scoring (L3)        &    1.5 &  0.1 & $<$0.01 \\
\diff{MD} graph (L4--5)        &    6.8 &  0.3 & 0.34 \\
\diff{MD} diffusion (L6)       &    3.6 &  0.1 & 0.15 \\
\midrule
Total                          & 2590.0 &  & \\
\bottomrule
\end{tabular}
\par}
\captionof{table}{\textbf{Computational cost of \methodname{}}. Per-image inference cost
at $N=4096$, $d=3072$, with $T_{\mathrm{it}}{=}30$ and line numbers referring to
Algorithm~1. All stages are paid on every image; the one-off harvest of
$\bm{\Pi}$ (Eq.~6) is offline and excluded here. \ens{TDE} contributes no
arithmetic of its own, so its cost is the two additional capture passes charged
to the first row. Memory is the increment over the resident
ConceptAttention capture, which occupies most of the $36.9$\,GB peak.}
\label{supp:tab-cost}
\end{table}
 
The refinement itself is close to free. Prototype scoring, graph construction
and diffusion together take $11.9$\,ms, which is $0.46\%$ of the full method
and $1.4\%$ of a single capture pass. Graph construction dominates that budget
and is quadratic in the token count, so its share grows with resolution rather
than staying fixed, but at the grid we use it is $0.3\%$ of the total.

\paragraph{The ensemble is the cost.}
A single capture pass takes $859$\,ms, so the unrefined read-out at $\tau_0$
runs in that time. \methodname{} takes $2590$\,ms, or $3.01\times$ as long, and
essentially all of the difference is the two additional capture passes that
\ens{TDE} requires. Reporting only the $0.46\%$ figure would therefore
understate the method by a factor of three. Set against the mean contributions
in Table~\ref{tab:ablation}, the two arithmetic operators return $3.7$\,ms per
mIoU point while the ensemble returns $1228$\,ms per point, a ratio of over
$300$. A deployment constrained by latency should keep \diff{manifold
diffusion} and the \prot{prototype channel} and set $|\mathcal{T}|=1$, which
retains $74\%$ of the gain over the base capture (row~6 of
Table~\ref{tab:ablation}) at a third of the latency.

\paragraph{Offline cost.}
Harvesting is $|\mathcal{U}|\times|\mathcal{T}|$ forward passes, $900$ at the
setting we use, or roughly $13$ minutes, paid once per vocabulary. It does not
affect inference, since the bank is $\bm{\Pi}\in\mathbb{R}^{K\times d}$
regardless of how many images built it, so a larger pool costs offline time and
nothing else.
 
\section{Implementation Details}
\label{supp:impl}
\paragraph{Concepts and background.}
Concept anchors are the dataset class names. On the two benchmarks that score an
explicit background class, background is represented not by one word but by a
fixed set of seven generic scene concepts, \emph{background}, \emph{floor},
\emph{grass}, \emph{tree}, \emph{sky}, \emph{wall} and \emph{road}, whose
columns are collapsed to the background label at read-out; this is what makes
the background column of $\mathbf{S}$ competitive with the foreground ones.
Background is excluded from the prototype bank, as noted in
Appendix~\ref{supp:datasets}.

\paragraph{Evaluation protocol.}
The read-out is a per-token argmax over concept columns, upsampled to the
evaluation grid. Two protocols appear, matched to how each benchmark is scored
in prior work, and the distinction is load-bearing: it moves the numbers
substantially and a mismatch is the most likely cause of a failed reproduction.
On PASCAL VOC the score field is built per image over the classes present in
that image plus the seven background concepts, following the ConceptAttention
VOC protocol, and this is what our baseline reproduces. On ADE20K,
Pascal-Context, Cityscapes and the two COCO benchmarks the read-out is taken
over the full vocabulary for every image, with per-image mAP over the
ground-truth-present classes and void pixels ignored. Both are implemented in
the released evaluators. Results use a single seed ($4$).

\paragraph{Reproducibility notes.}
FLUX is run with offloading disabled so that every sub-module resolves to one
device. The released configuration dataclass ships conservative defaults
($\alpha = 0.85$, $\kappa = 15$, $\rho = 0$, the bare capture baseline); the
paper settings are those passed on the command line and reproduced in the
README, and it is those that every reported number uses. The main environment is
Python $3.10$ with PyTorch $\ge 2.4$, diffusers $\ge 0.35$ and transformers
$\ge 4.44$; the Seg4Diff experiments run in a separate environment against that
method's pinned diffusers fork and \texttt{transformers==4.44.2}, into which the
backbone-independent MaViSeg operators load without modification.
\section{Capture Agnosticity of \methodname{}}
\label{supp:agnosticity}

Does \methodname{} necessarily need a diffusion transformer to refine? We argue
it does not. The refinement operator of Eq.~(4) sees only the pair
$(\mathbf{S},\mathbf{X})$ and never queries the model that produced it, so what
the method needs is a property of the capture rather than of the architecture.
It needs the score field and the token features to share a common set of $N$
positions. It needs $\mathbf{X}$ to place instances of a class nearer one
another than to instances of another class, since that is the geometry
\prot{BVP} harvests and \diff{MD} propagates over. And it needs the scorer to
be reliable enough to surface confident exemplars of each concept, which is a
weaker demand than asking it to be calibrated across concepts. Diffusion, joint
attention and a generative objective appear nowhere in that list.

The baselines of Table~\ref{tab:main} mostly satisfy it already. A U-Net
capture offers per-concept cross-attention as $\mathbf{S}$ and decoder
activations as $\mathbf{X}$; a discriminative read-out offers a relevance map
over patch tokens alongside the tokens themselves. The U-Net case does ask for
two accommodations that MM-DiT does not, namely that captures taken at several
resolutions be brought to a common grid, and that one decide which block
supplies $\mathbf{X}$ where MM-DiT presents a single token stream. The
normalizer $g(\cdot)$ needs no such care, defined as it is by a median decision
margin rather than by the scale of the scores beneath it.

\ens{TDE} is the operator that does not travel. It presumes a trajectory to
average over, so a backbone without an iterative sampling process leaves it
nothing to ensemble and the stack falls back to \prot{BVP} $+$ \diff{MD}, the
configuration of row~7 in Table~\ref{tab:ablation}. That row still improves on
the capture, so the reduction is graceful rather than fatal, and the two
operators that survive it are also the two that cost almost nothing
(\S\ref{supp:cost}). If anything we would expect the U-Net captures to gain
more than the MM-DiT ones, since the mAP--mIoU dissociation they show in
Table~\ref{tab:main} is the cross-concept calibration failure
\prot{BVP} was built for, and since refinement has helped most wherever the raw
capture is weakest.